\documentclass[]{spie}  

\usepackage{amsmath,amsfonts,amssymb}
\usepackage{graphicx}
\usepackage[colorlinks=true, allcolors=blue]{hyperref}
\usepackage{subcaption}
\usepackage{multirow}
\usepackage{xcolor}
\usepackage{pifont}
\newcommand{\cmark}{\ding{51}}%
\newcommand{\xmark}{\ding{55}}%

\title{Utilizing Grounded SAM for self-supervised frugal camouflaged human detection}

\author[a,b]{Matthias Pijarowski}
\author[a]{Alexander Wolpert}
\author[b]{Martin Heckmann}
\author[a]{Michael Teutsch}
\affil[a]{Hensoldt Optronics GmbH, Oberkochen, Germany}
\affil[b]{Aalen University of Applied Sciences, Germany}

\authorinfo{Contact: \{matthias.pijarowski, alexander.wolpert, michael.teutsch\}@hensoldt.net, martin.heckmann@hs-aalen.de}

\begin{document} 
\maketitle

\begin{abstract}
Visually detecting camouflaged objects is a hard problem for both humans and computer vision algorithms. Strong similarities between object and background appearance make the task significantly more challenging than traditional object detection or segmentation tasks. Current state-of-the-art models use either convolutional neural networks or vision transformers as feature extractors. They are trained in a fully supervised manner and thus need a large amount of labeled training data. In this paper, both self-supervised and frugal learning methods are introduced to the task of Camouflaged Object Detection (COD). The overall goal is to fine-tune two COD reference methods, namely SINet-V2 and HitNet, pre-trained for camouflaged animal detection to the task of camouflaged human detection. Therefore, we use the public dataset CPD1K that contains camouflaged humans in a forest environment. We create a strong baseline using supervised frugal transfer learning for the fine-tuning task. Then, we analyze three pseudo-labeling approaches to perform the fine-tuning task in a self-supervised manner. Our experiments show that we achieve similar performance by pure self-supervision compared to fully supervised frugal learning.  
\end{abstract}


\section{INTRODUCTION}
\label{sec:intro}
Camouflaged Object Detection (COD) aims at segmenting camouflaged objects, i.e. objects blending into the background by either natural or artificial camouflage~\cite{Fan.b}. Camouflaged object detection has many real world applications such as medical segmentation, where it can help experts locate polyps~\cite{Fan.6132020}, rare species discovery~\cite{LaPerezdeFuente.2012}, and search \& rescue missions, where it can help finding survivors for example in snowy regions~\cite{Rizk.}. COD is usually defined as a binary segmentation task~\cite{Lv2023}. The current state-of-the-art performance is achieved by deep learning-based approaches utilizing Convolutional Neural Networks (CNNs)~\cite{Fan.b} or Transformers~\cite{Hu.3222022} together with a fully supervised training strategy. Hence, they need a large amount of labeled training data. Annotation, however, is expensive since each pixel should be labeled in a segmentation task~\cite{Reiss2021} and precise localization of a camouflaged object in a scene is obviously difficult for a human observer~\cite{Lamdouar2023}. Labeling a single image can take up to one hour~\cite{Fan.b}. This is why various authors aim at reducing the annotation task for image segmentation approaches~\cite{Cheng2022,Reiss2021,Li2023,Papadopoulos2021}. Modern foundation models such as \emph{Segment Anything}~\cite{Kirillov.452023} together with meaningful text prompts provide powerful zero-shot performance in image segmentation~\cite{Luddecke2022,Zhou2023,Ma2024}. This applies even to the more challenging COD task of segmenting camouflaged animals~\cite{He2023,Hu2024}.

In this paper, we tackle the challenge of both self-supervised and frugal learning for the detection of camouflaged humans. Within COD, camouflaged human detection is a niche topic and thus it has been rather little researched~\cite{Zheng.2019,Liu2023}. We utilize the public CPD1K dataset~\cite{Zheng.2019} and adopt state-of-the-art COD methods that detect camouflaged animals~\cite{Fan.b,Hu.3222022}. Those methods are fine-tuned with the entire CPD1K training dataset in a fully supervised way to the task of camouflaged human detection as a baseline. Then, we show that we can achieve nearly similar performance by using an unlabeled fraction of the CPD1K training data by utilizing foundation model-based pseudo-labeling and $k$-shot frugal learning. Our contributions can be summarized as follows: (1)~we show that supervised frugal learning with a fraction of about 6\,\% of the full training data can achieve good performance for camouflaged human detection. (2)~we demonstrate that Grounded SAM (GSAM)~\cite{ren2024grounded} can be a powerful tool to generate pseudo-labels for a self-supervised learning scheme. (3)~ in our experiments, we achieve a relative performance gap of only about 10\,\% between the fully supervised model trained on the full training dataset compared to our self-supervised frugally learning model for camouflaged human detection

The remainder of this paper is organized as follows: related work is reviewed in Section~\ref{sec:relwork}. The methodology for both frugal and self-supervised learning for the COD task is described in Section~\ref{sec:methods}. Comprehensive experiments are conducted and discussed in Section~\ref{sec:experiments}. Conclusions are given in Section~\ref{sec:conclusion}.

\section{RELATED WORK}
\label{sec:relwork}

\textbf{Camouflaged object detection:} COD aims at detecting camouflaged objects. In the literature, this task is usually formulated as a binary segmentation problem~\cite{Fan.b}. Due to high similarities between object and background, COD is usually more challenging than traditional object segmentation tasks. Early approaches to COD used different handcrafted features like texture and contrast~\cite{Huerta.}, 3d-convexity~\cite{Tankus.}, optical flow \cite{jianqin.2011} or try to solve the problem in the frequency domain~\cite{Li_2018}. As the complexity of the scene rises, these approaches often struggle to segment the camouflaged objects. Recently deep learning models have been utilized to solve the COD task~\cite{Fan.2020}. Deep learning models have shown that they are able to learn powerful representations from data~\cite{Mei2021,Hu.3222022,Fan.b}. Some approaches solve the COD task by utilizing auxiliary tasks such as classification~\cite{Le.2019}, object ranking~\cite{Lv.362021} or boundary detection~\cite{sun.2022}. Other approaches try to mimic the behaviour of predators such as search and identification~\cite{Fan.b} or zooming in and out~\cite{pang2022zoom}. Vision Transformers~\cite{Dosovitskiy.10222020} have also found application in COD. Using transformer-based feature extractors has helped surpass the performance of Convolutional Neural Network (CNN) based architectures in many cases~\cite{Hu.3222022,huang2023feature,yin2022camoformer}. Although deep learning based models have shown great results they have a drawback: in order to achieve satisfying performance they need large amounts of labeled data, which is often not easy to acquire and thus motivates us to look into self-supervised and frugal learning methods for the COD task.

\noindent\textbf{Self-supervised learning:} Self-supervised learning (SSL) focuses on learning from data without labels. Either the data can be used directly by masking image regions and then predicting the masked regions~\cite{devlin-etal-2019-bert,he2021masked} or by generating pseudo-labels, which can then be used as labels during training~\cite{LOST,wang2021comining,Li2020}. In this paper, we focus on the popular pseudo labeling technique, where labels for unlabeled data are generated through some model and subsequently used as Ground Truth (GT) labels. Some approaches \cite{lee2013pseudolanel,sohn2020fixmatch} use a single model generating new pseudo-labels each epoch and getting re-trained on its own pseudo-labels. Student-Teacher frameworks~\cite{xu2021softteacher,xu2023efficient,wang2023consistentteacher} use two models: a teacher to generate pseudo-labels and a student, who is trained uses those labels. The teacher is updated through an Exponential Moving Average (EMA) to improve the pseudo labels. Other approaches use an auxiliary model that is not updated during training to generate the pseudo-labels~\cite{wang2022openworld}. Since the first two methods require re-training of the pseudo-label generator, they are not easily adaptable to COD. Thus we focus on approaches utilizing a frozen auxiliary model: inpainting and segmentation foundation models.\\
Image inpainting is the task of reconstructing missing regions in an image with plausible contents~\cite{Zeng.5242020}. This task has many applications in imaging and graphics applications, e.g. object removal, image restoration, etc. Early inpainting methods use image statistics to take patches from the image itself or similar images and paste those over the missing regions \cite{Barnes:2009:PAR}. These methods lack structural understanding of the image and are limited to existing images. Recent methods utilize deep learning. They use encoder decoder architectures, which learn to predict masked regions of the input image \cite{pathak2016context,yang2017highresolution,song2018contextualbased}. These methods can generate new contents based on the learned parameters. Mask-Aware Transformer (MAT)~\cite{Li.3292022} is the current state-of-the-art model for image inpainting on multiple datasets. It was trained on the places dataset, which contains more than 10 million images and a wide range of diverse scenes. In contrast to early methods as well as most CNN based methods it is capable of large hole inpainting, due to the transformer based architecture, which is able to model large range dependencies between pixels. Large hole refers to the ratio of the mask used to cover parts of the image compared to the image.\\
A foundation model is a model trained on broad data, most commonly in a self-supervised way and it can be adapted to a wide range of downstream tasks~\cite{Bommasani2021FoundationModels}. Google's BERT \cite{devlin-etal-2019-bert} and OpenAI's GPT \cite{radford2018gpt} series are examples for foundation models in the language domain. In the computer vision domain, Contrastive Language-Image Pretraining (CLIP)~\cite{radford2021learning} is a popular foundation model, which predicts image and text pairings and which is widely used as basis for downstream tasks~\cite{Lin2022,Rasheed2023}. Segment Anything (SAM)~\cite{Kirillov.452023} is a recently introduced segmentation foundation model. Given an image and a point, bounding box, polygon or text\footnote{Not available in the public version of SAM} as additional input SAM segments at least one object in the proposed region. Grounding DINO~\cite{Liu.392023} is an object detector based on DETR with Improved DeNoising Anchor Boxes for End-to-End Object Detection (DINO)~\cite{Zhang.372022}. It extends DINO by encoding a text phrase in addition to an image. The text phrase is then used to find objects on the image and generate bounding box proposals. It shows strong zero-shot performance on novel data. GSAM~\cite{ren2024grounded} combines Grounding DINO and SAM. First Grounding DINO is used to generate bounding box proposals. The bounding boxes are then used as input for SAM to generate a segmentation mask.

\noindent\textbf{Frugal learning:} Frugal learning emphasises the cost associated with the use of data and computational resources. Evchenko et al. \cite{evchenko2021frugal} categorize frugal machine learning in the following three categories: (1)~\emph{Input frugality} focuses on the cost associated with data. The cost can be that of the acquisition of the data, or the exploitation of its features. Frugal inputs can be the usage of less training samples or less of the data's features. Input frugality can be motivated by small amounts of available data. (2)~\emph{Learning process frugality} focuses on the cost of the learning process, especially computational resources. Frugal learning processes might produce a less accurate model than one trained under an optimal setting, but it uses the available computational resources more efficiently. Learning process frugality can be motivated by limited computational resource, e.g. limited processing power or limited battery capacity. (3)~\emph{Model frugality} focuses on the cost associated with storing or using a model. A frugal model might produce less accurate results, but uses less memory and produces the results with less computational effort. Model frugality is motivated by limited computational resources, e.g. the amount of memory available or limited processing power. As mentioned in Section~\ref{sec:intro} the COD task is mainly limited by the amount of available data. To alleviate the described data challenges of COD we focus on input frugality, especially frugality of data and labels.\\
A closely related task is Few-Shot Object Detection (FSOD). FSOD aims at learning from abundant object instances in a non-target domain and few new object instances in the target domain. Kang et al.~\cite{Kang.1252018} propose to split a dataset into base and novel classes with abundant base class samples and only few novel class samples. They then train a model on the base classes and fine-tune it on $k$ shots of the novel classes. A shot is defined as exactly one annotated object. But they only evaluate the model performance on novel classes. Wang et al.~\cite{Wang.3162020} argue that previous FSOD evaluation protocols suffer from statistical unreliability. They propose that (1)~the models should be evaluated on base and novel classes to detect performance decreases on those classes. Also (2)~multiple iterations $i$ of the models should be trained for each k since the sample variance could be huge, especially for low $k$ which might lead to the model performance being over-/underestimated depending on the samples selected for fine-tuning.


\section{METHODOLOGY}
\label{sec:methods}

Fan et al.~\cite{Fan.b} define camouflaged object detection as a class-agnostic segmentation task. Given an image, the camouflaged object detection algorithm has to assign each pixel $i$ a binary label $\hat y_i \in \{0, 1\}$. The algorithm should predict 1 when the pixel is part of the camouflaged object and 0 otherwise. We adopt this problem formulation when tackling the task of COD in this paper. In this section, we first set a strong baseline for fully supervised frugal learning in COD using the (to the best of our knowledge) only publicly available dataset for camouflaged human detection: CPD1K~\cite{Zheng.2019}. Then, we explore methods for generating pseudo-labels. In this way, we approach the topic of self-supervised learning. The combination of both frugal and self-supervised learning is then analyzed in the experiments in Section~\ref{sec:experiments}.

\subsection{Frugal learning for COD}
\label{subsec:camouflaged_frugal_learning}

Based on COD and the FSOD evaluation protocol introduced in Section~\ref{sec:relwork}, we establish the following protocol for COD with limited training data: we first sample $k$ shots from a training dataset. Then, we fine-tune a base model on the aforementioned $k$ shots. Afterwards, we evaluate the fine-tuned model on standard COD evaluations measures using a test dataset. We repeat this process $i$ times for each $k$. We use 48\,\% of the data as train split (CPD1K train) for fine-tuning, 12\,\% of the data for validation, and the remaining 40\,\% were held out and used as test split (CPD1K test). We split the dataset once and then sample from the resulting splits. The splits were chosen according to Zheng et al.~\cite{Zheng.2019}. In the CPD1K dataset, an image contains exactly one object and thus one image is one shot. We use $k = \{1, 2, 3, 5, 10, 30, 50\}$ and $i = 30$.
The whole process can be summed up as: 
\begin{enumerate}
    \item Randomly sample $k$ images from the CPD1K train dataset.
    \item Fine-tune the base model on those $k$ samples.
    \item Evaluate the resulting model based on standard COD evaluation metrics.
    \item Repeat steps 1-3 $i$ times for each $k$.
\end{enumerate}
The results are reported and discussed in Section~\ref{subsec:frugalexperiments}.

\subsection{Generating pseudo-labels for self-supervised learning}
We will discuss three different approaches to generate pseudo-labels for the data. The methods utilized are: (1)~GSAM: a foundation model for image segmentation~\cite{ren2024grounded}, (2)~MAT: a state of the art model for image inpainting~\cite{Li.3292022}, and (3)~HitNet: a state-of-the-art model for camouflaged object detection~\cite{Hu.3222022}.

\begin{figure}
    \centering
    \begin{subfigure}{.16\textwidth}
        \centering
        \includegraphics[width=\textwidth]{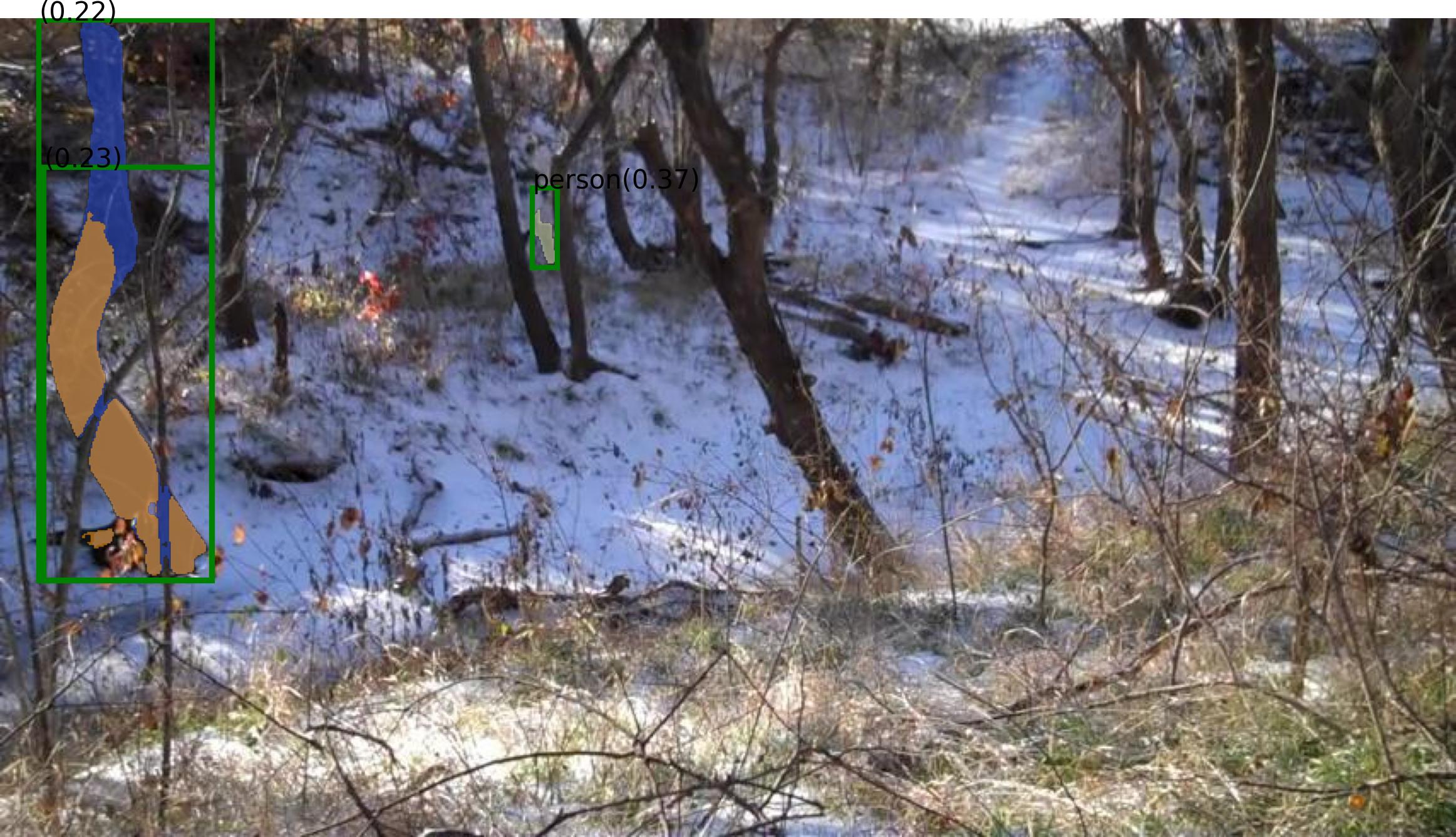}
    \end{subfigure}
    \begin{subfigure}{.16\textwidth}
        \centering
        \includegraphics[width=\textwidth]{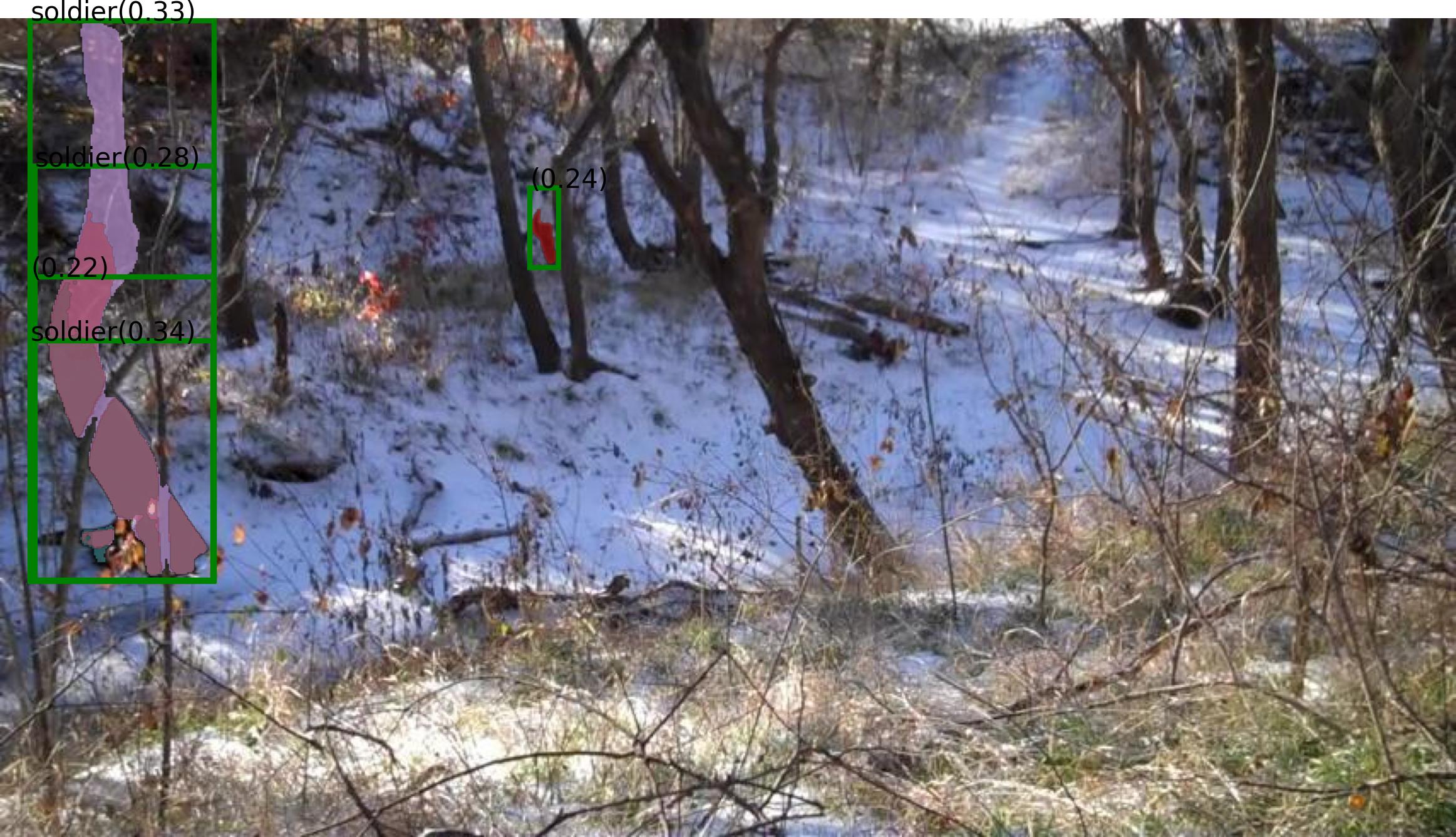}
    \end{subfigure}
    \begin{subfigure}{.16\textwidth}
        \centering
        \includegraphics[width=\textwidth]{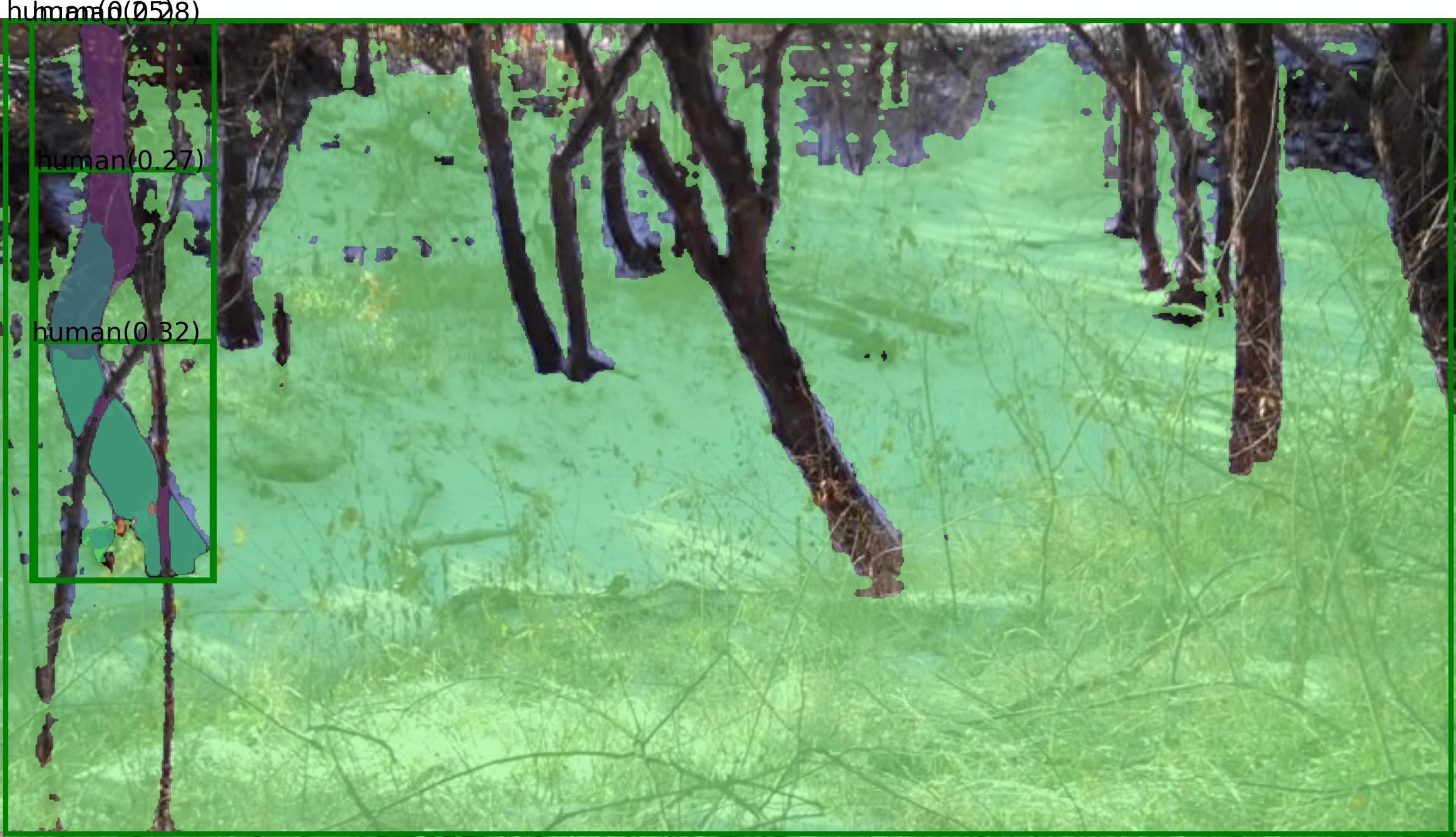}
    \end{subfigure}
    \begin{subfigure}{.16\textwidth}
        \centering
        \includegraphics[width=\textwidth]{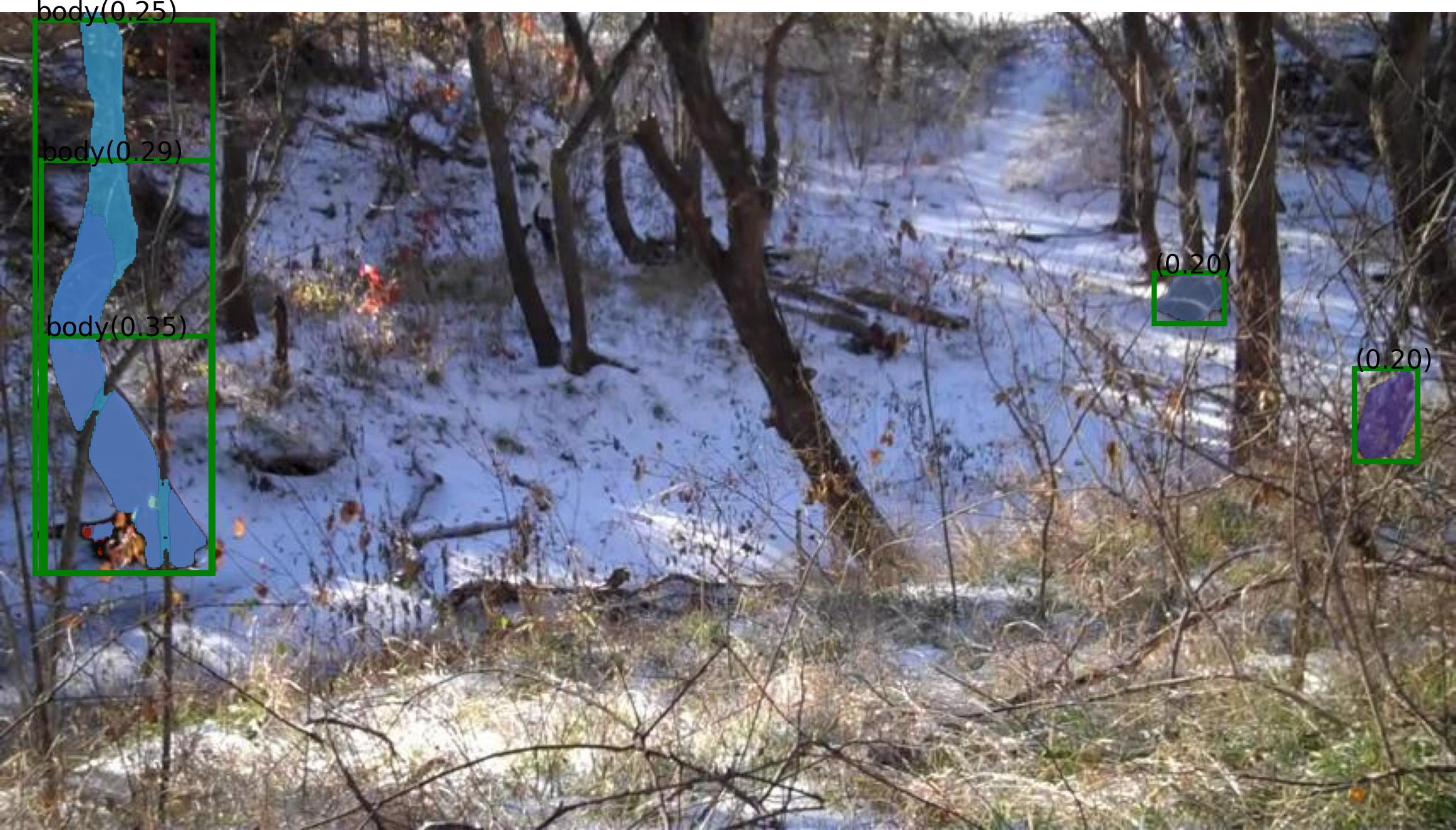}
    \end{subfigure}
    \begin{subfigure}{.16\textwidth}
        \centering
        \includegraphics[width=\textwidth]{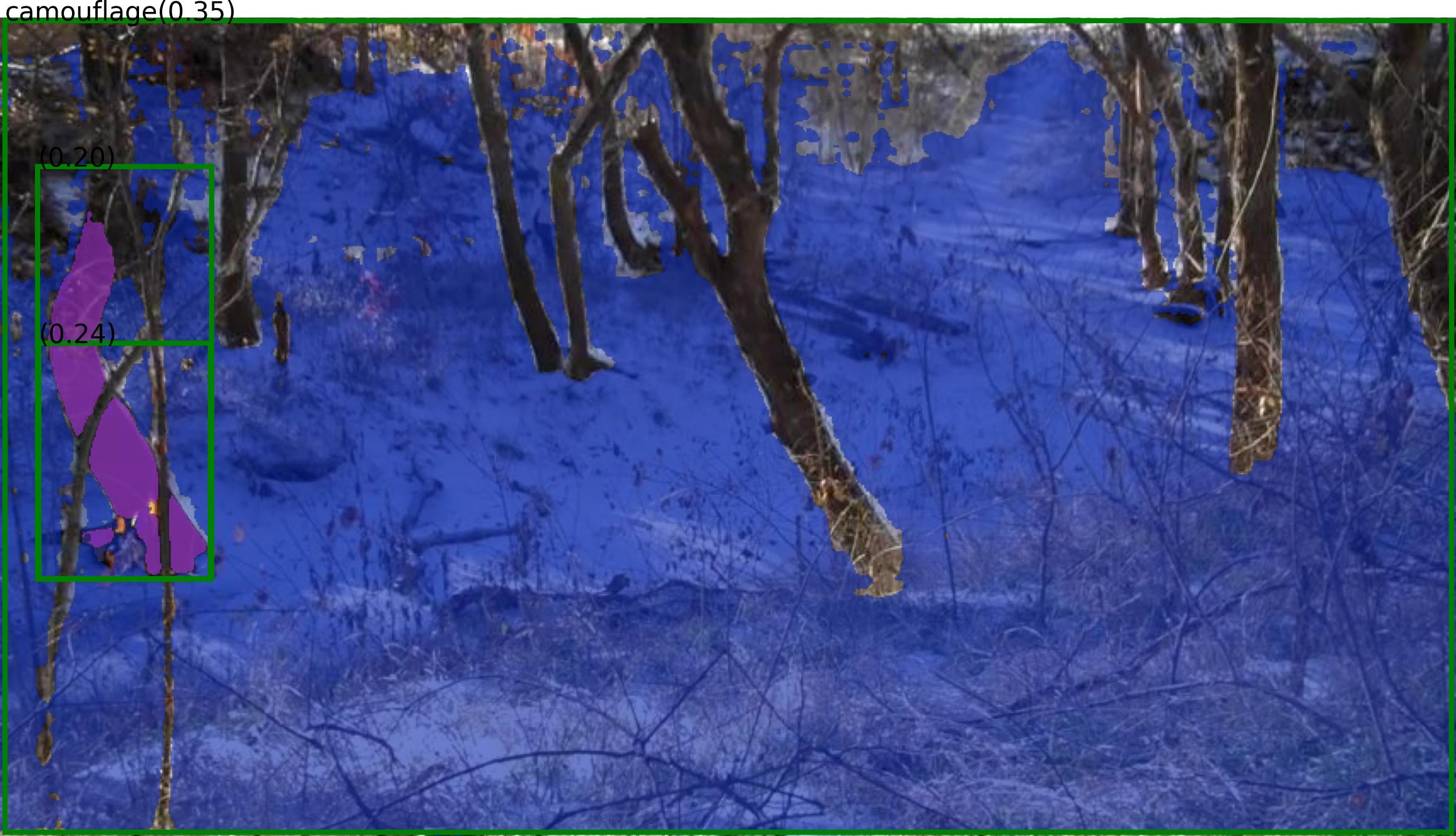}
    \end{subfigure}
    \begin{subfigure}{.16\textwidth}
        \centering
        \includegraphics[width=\textwidth]{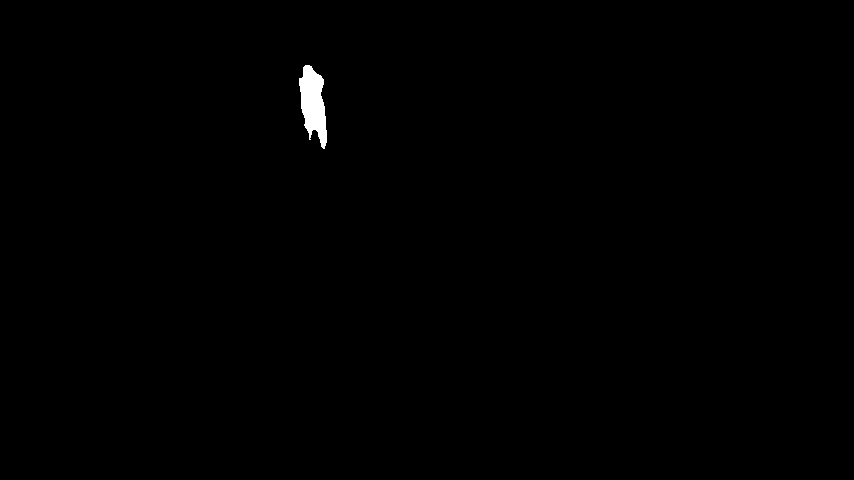}
    \end{subfigure}

    \begin{subfigure}{.16\textwidth}
        \centering
        \includegraphics[width=\textwidth]{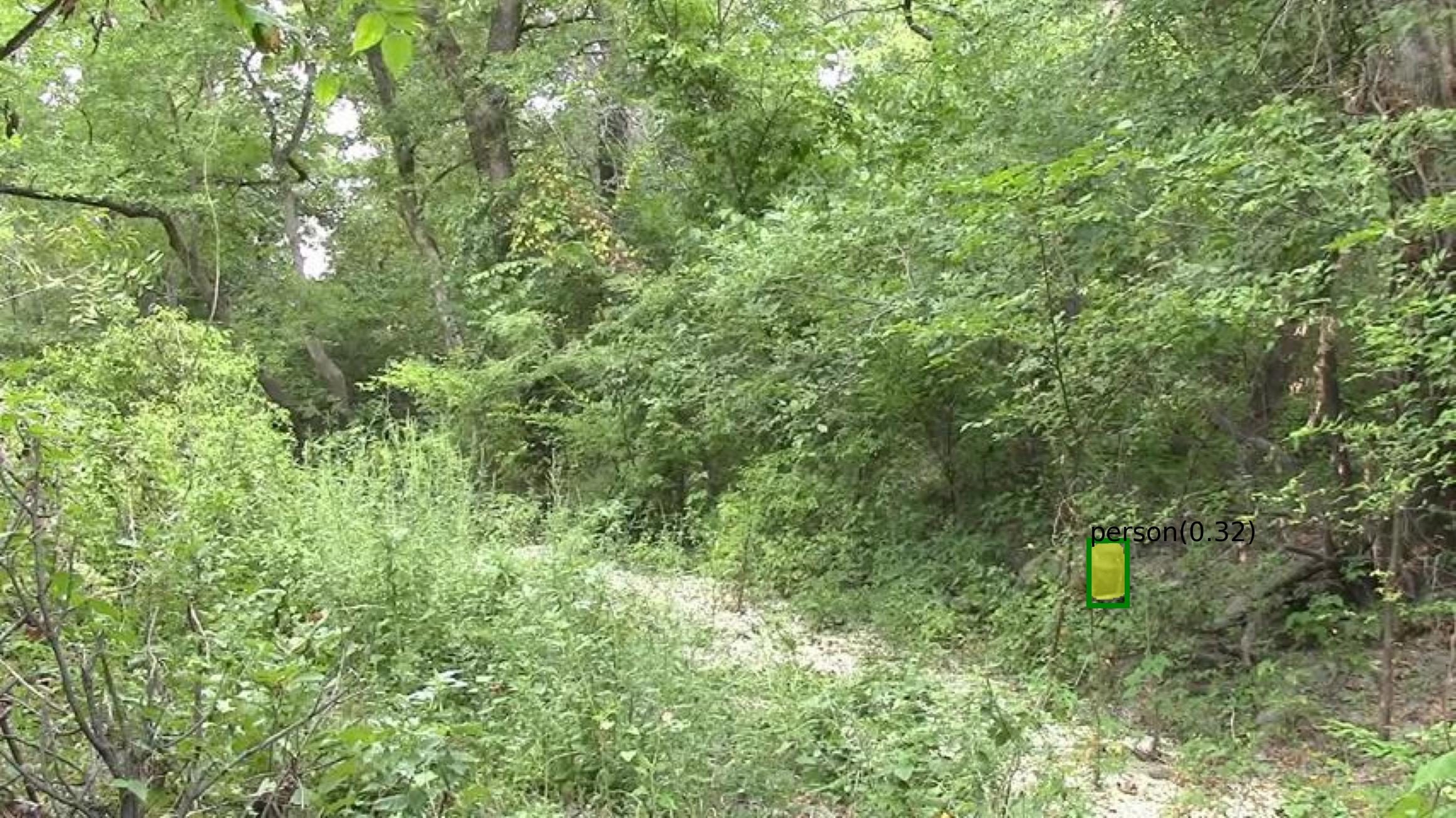}
    \end{subfigure}
    \begin{subfigure}{.16\textwidth}
        \centering
        \includegraphics[width=\textwidth]{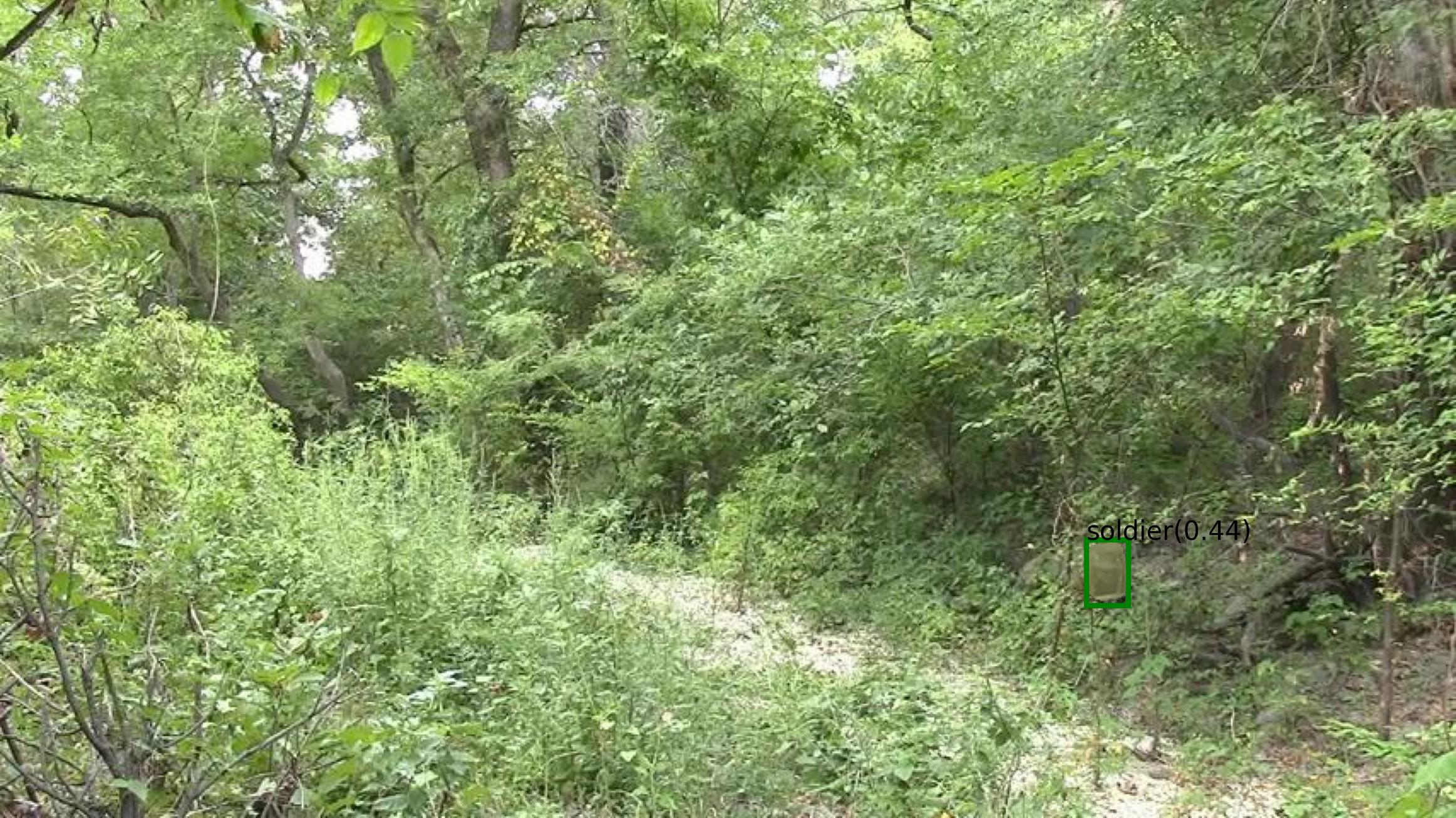}
    \end{subfigure}
    \begin{subfigure}{.16\textwidth}
        \centering
        \includegraphics[width=\textwidth]{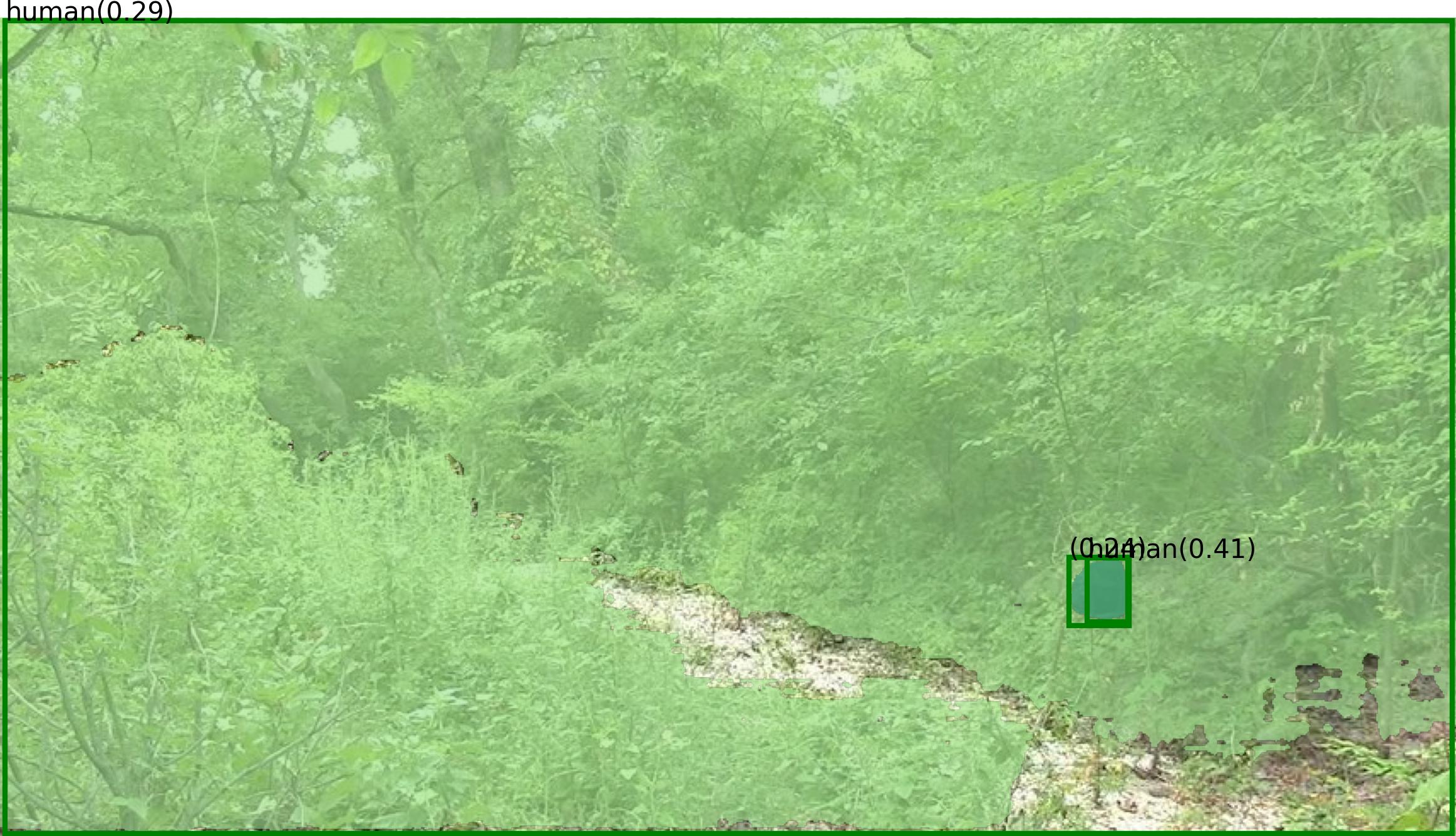}
    \end{subfigure}
    \begin{subfigure}{.16\textwidth}
        \centering
        \includegraphics[width=\textwidth]{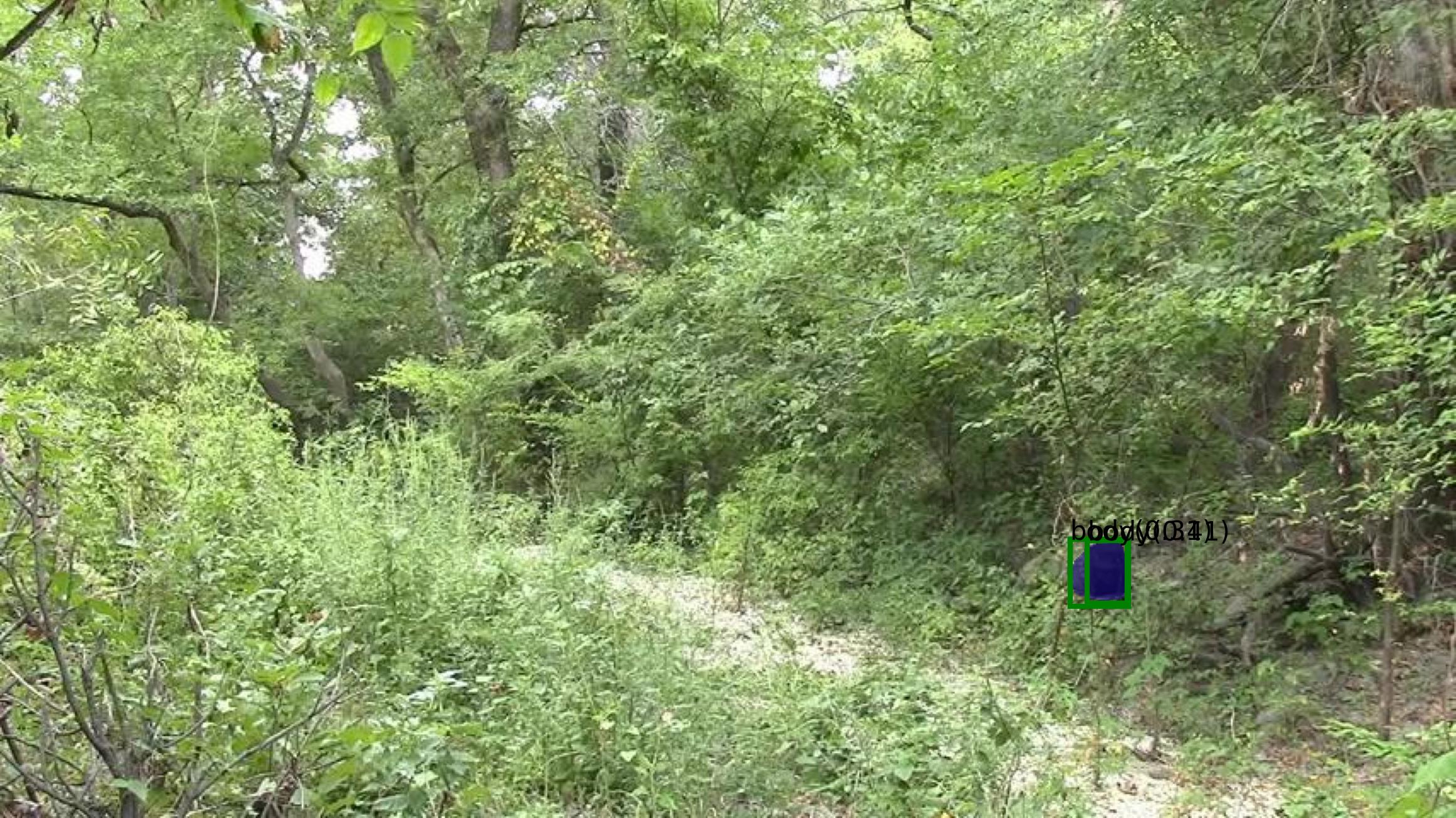}
    \end{subfigure}
    \begin{subfigure}{.16\textwidth}
        \centering
        \includegraphics[width=\textwidth]{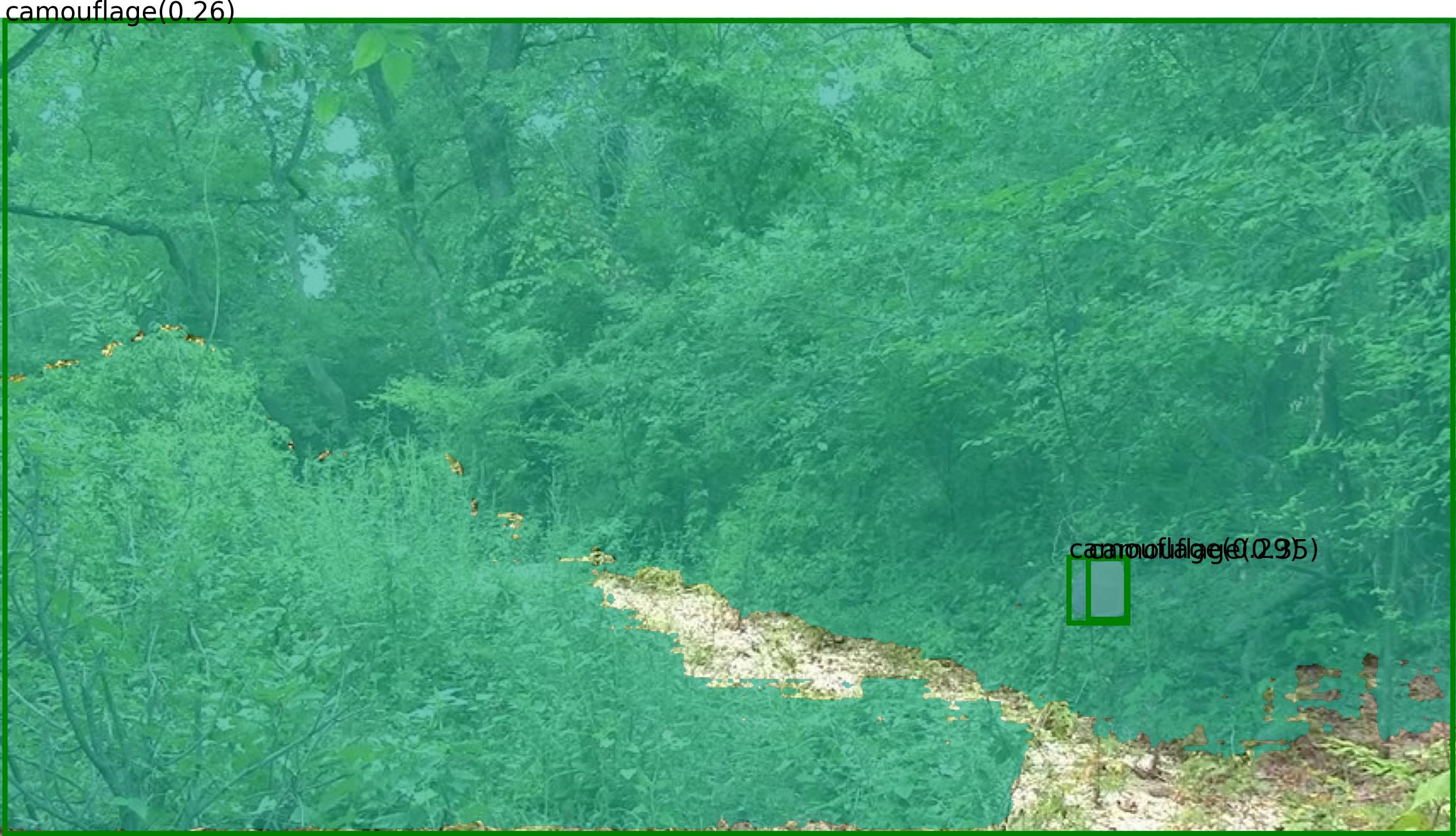}
    \end{subfigure}
    \begin{subfigure}{.16\textwidth}
        \centering
        \includegraphics[width=\textwidth]{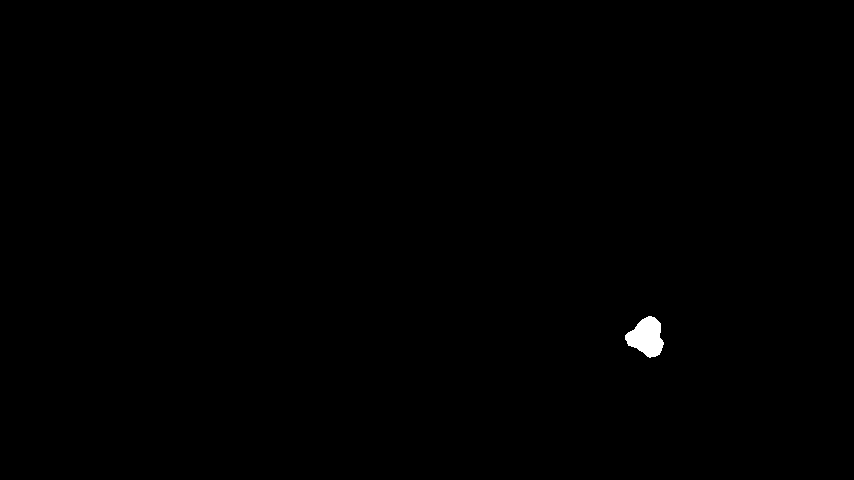}
    \end{subfigure}

    \begin{subfigure}{.16\textwidth}
        \centering
        \includegraphics[width=\textwidth]{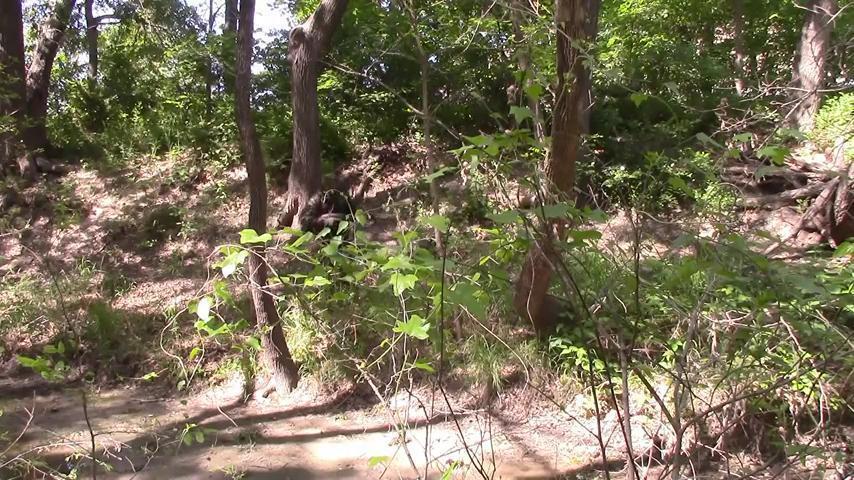}
    \end{subfigure}
    \begin{subfigure}{.16\textwidth}
        \centering
        \includegraphics[width=\textwidth]{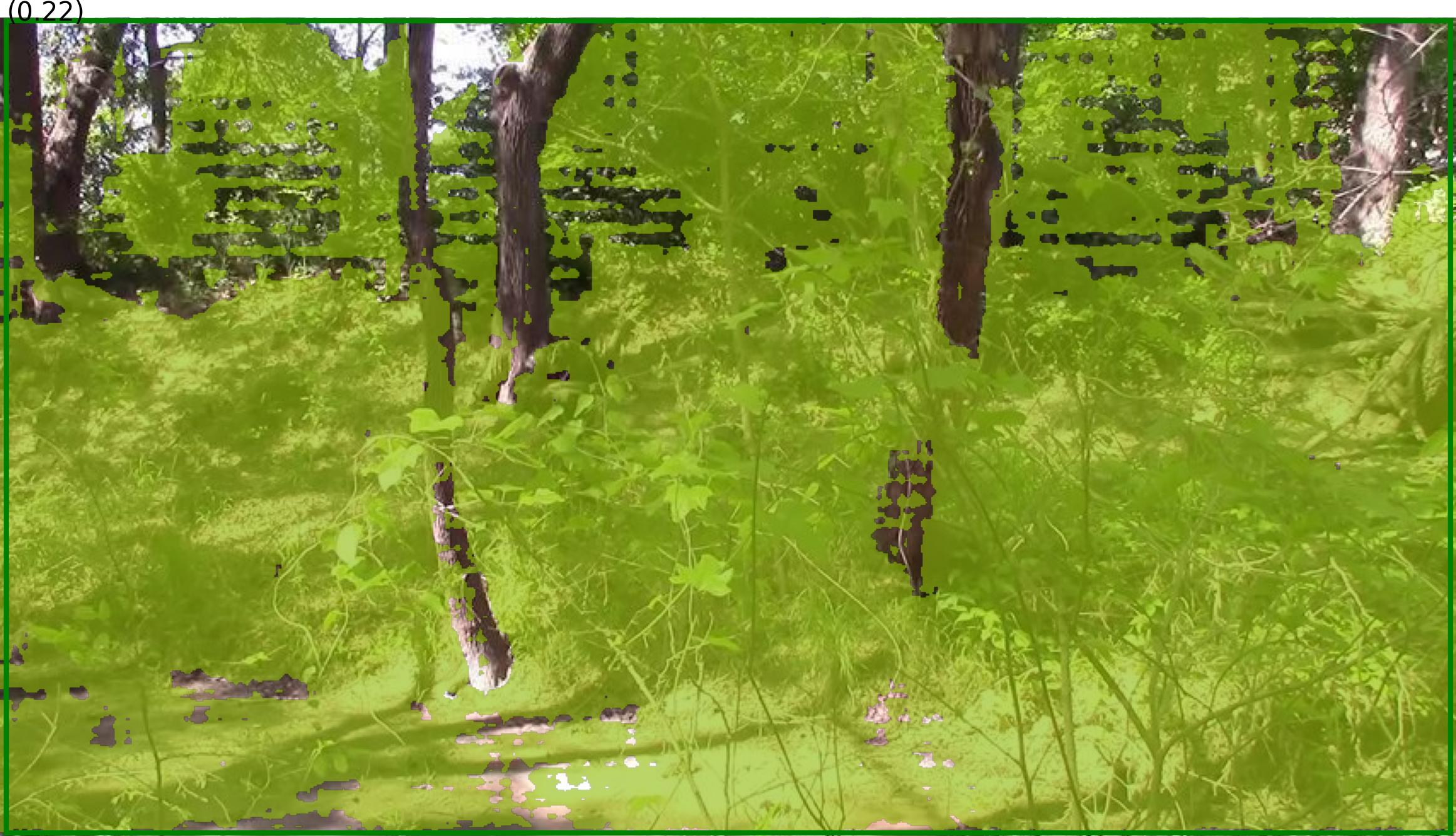}
    \end{subfigure}
    \begin{subfigure}{.16\textwidth}
        \centering
        \includegraphics[width=\textwidth]{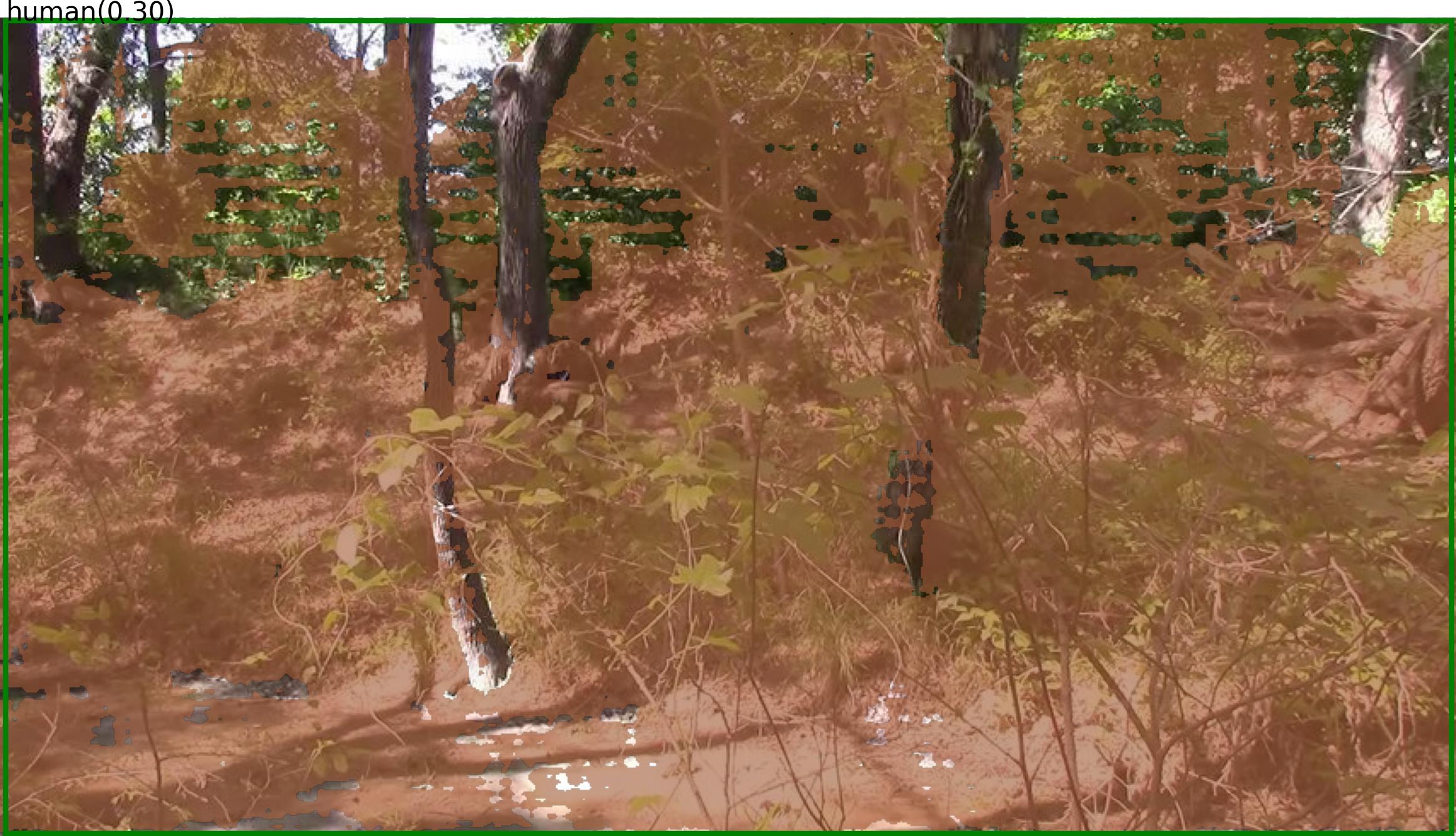}
    \end{subfigure}
    \begin{subfigure}{.16\textwidth}
        \centering
        \includegraphics[width=\textwidth]{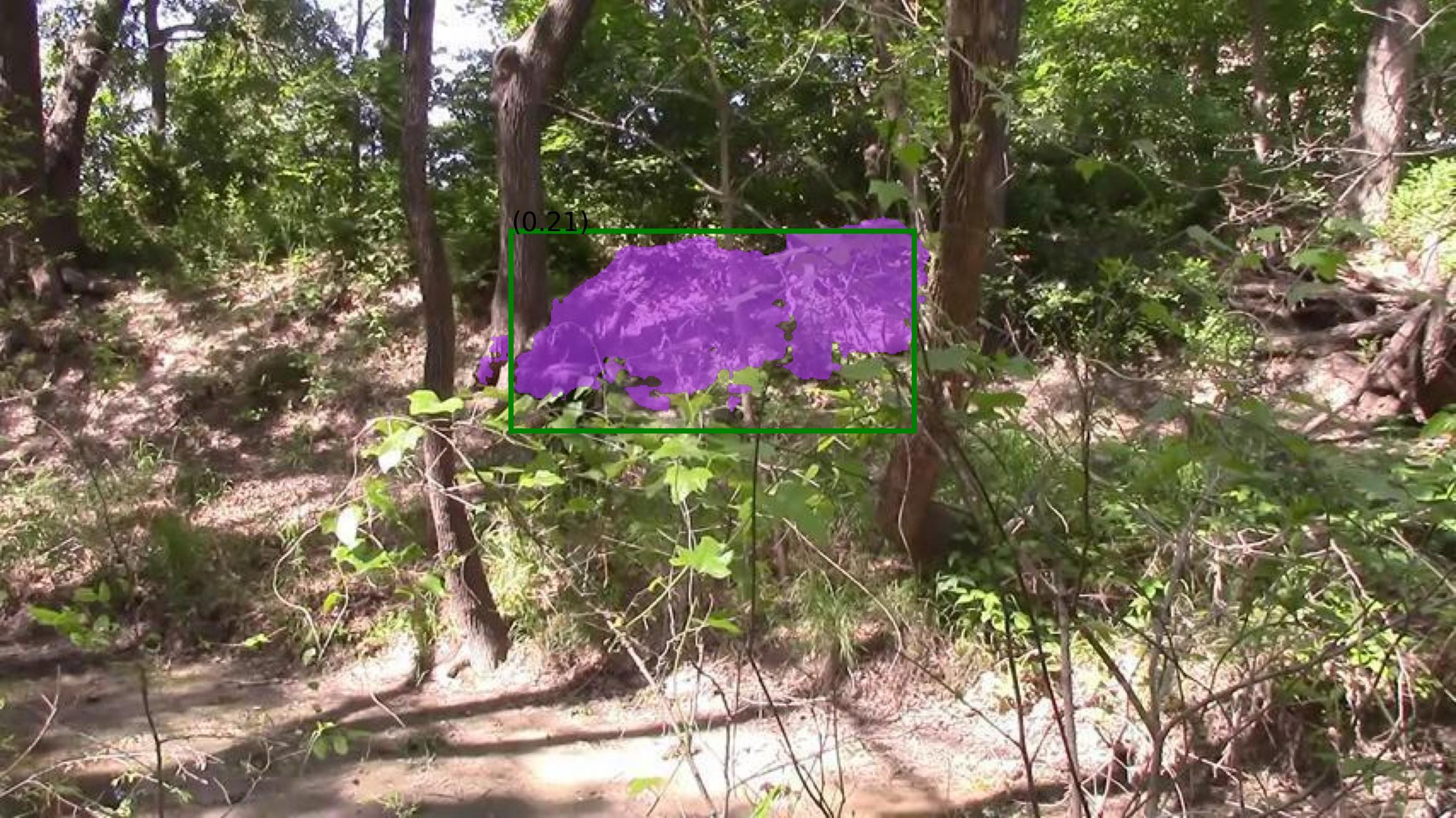}
    \end{subfigure}
    \begin{subfigure}{.16\textwidth}
        \centering
        \includegraphics[width=\textwidth]{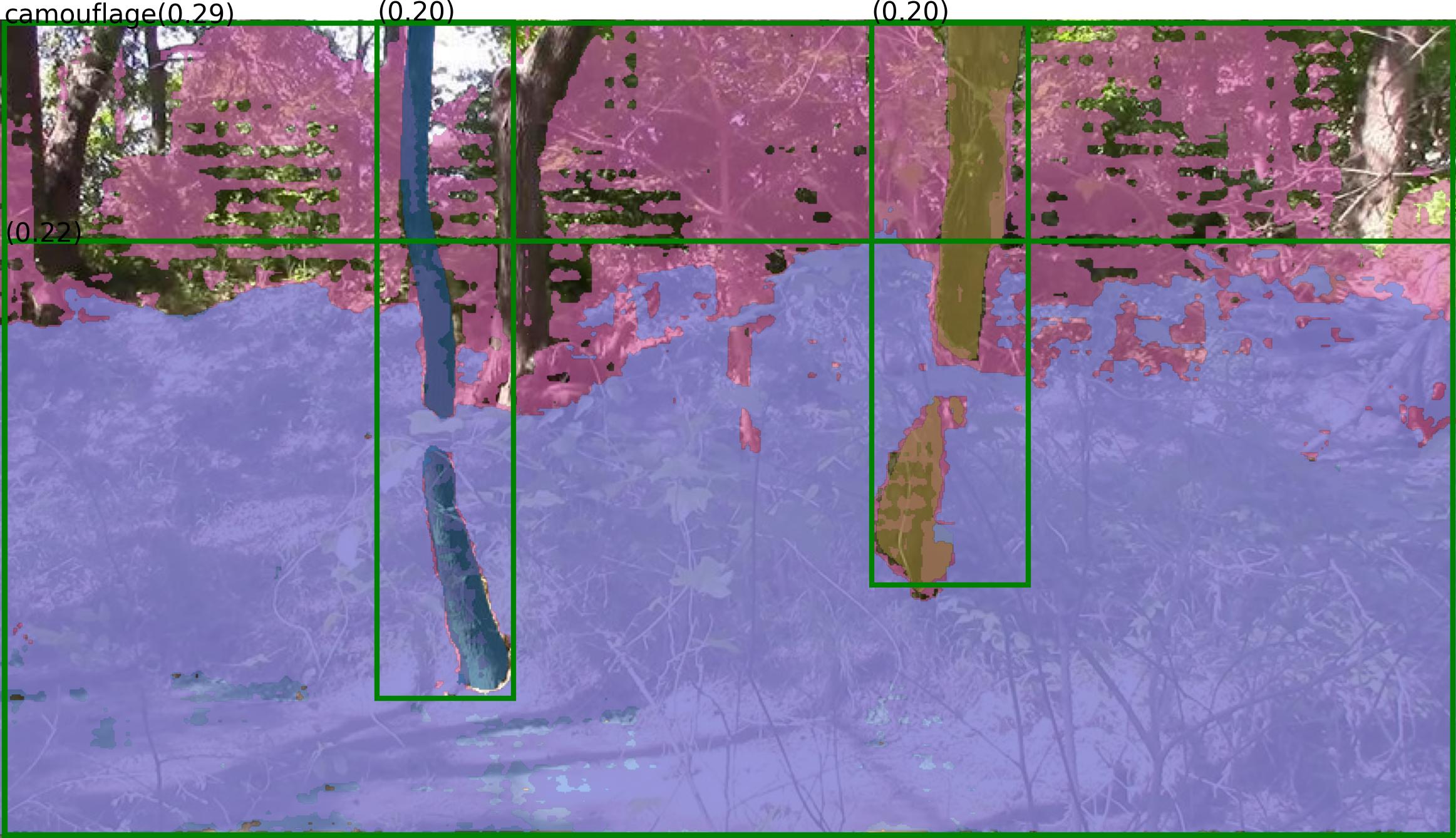}
    \end{subfigure}
    \begin{subfigure}{.16\textwidth}
        \centering
        \includegraphics[width=\textwidth]{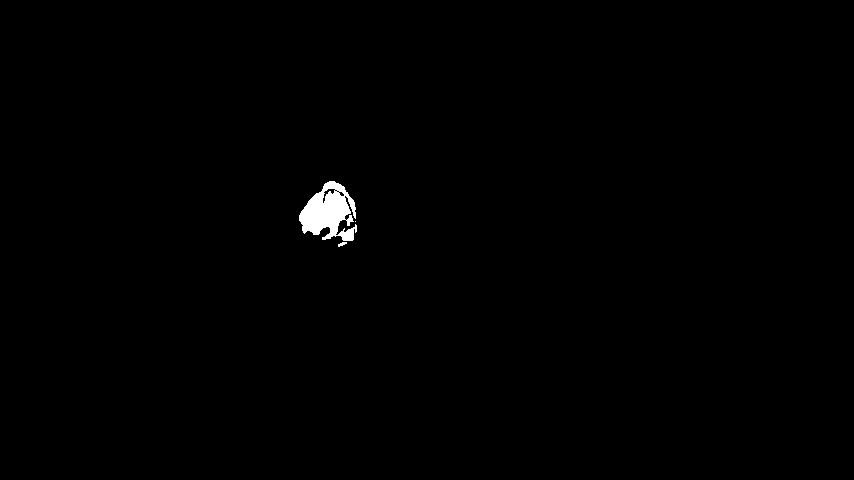}
    \end{subfigure}

    \begin{subfigure}{.16\textwidth}
        \centering
        \includegraphics[width=\textwidth]{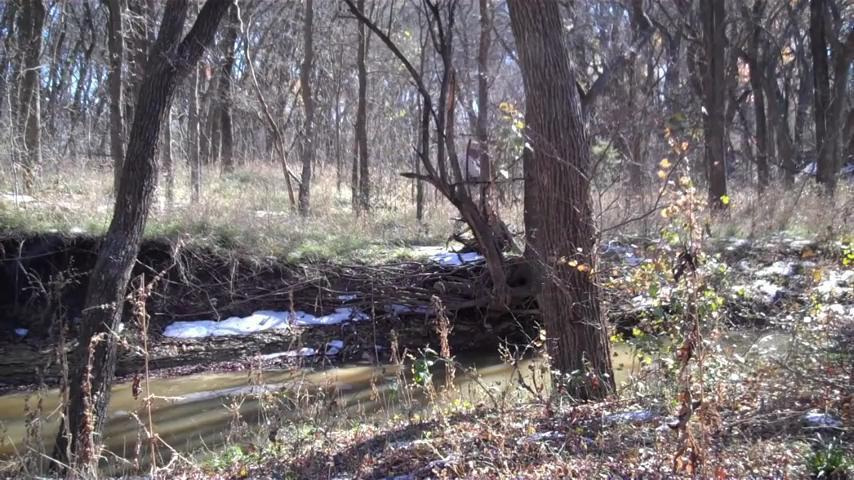}
    \end{subfigure}
    \begin{subfigure}{.16\textwidth}
        \centering
        \includegraphics[width=\textwidth]{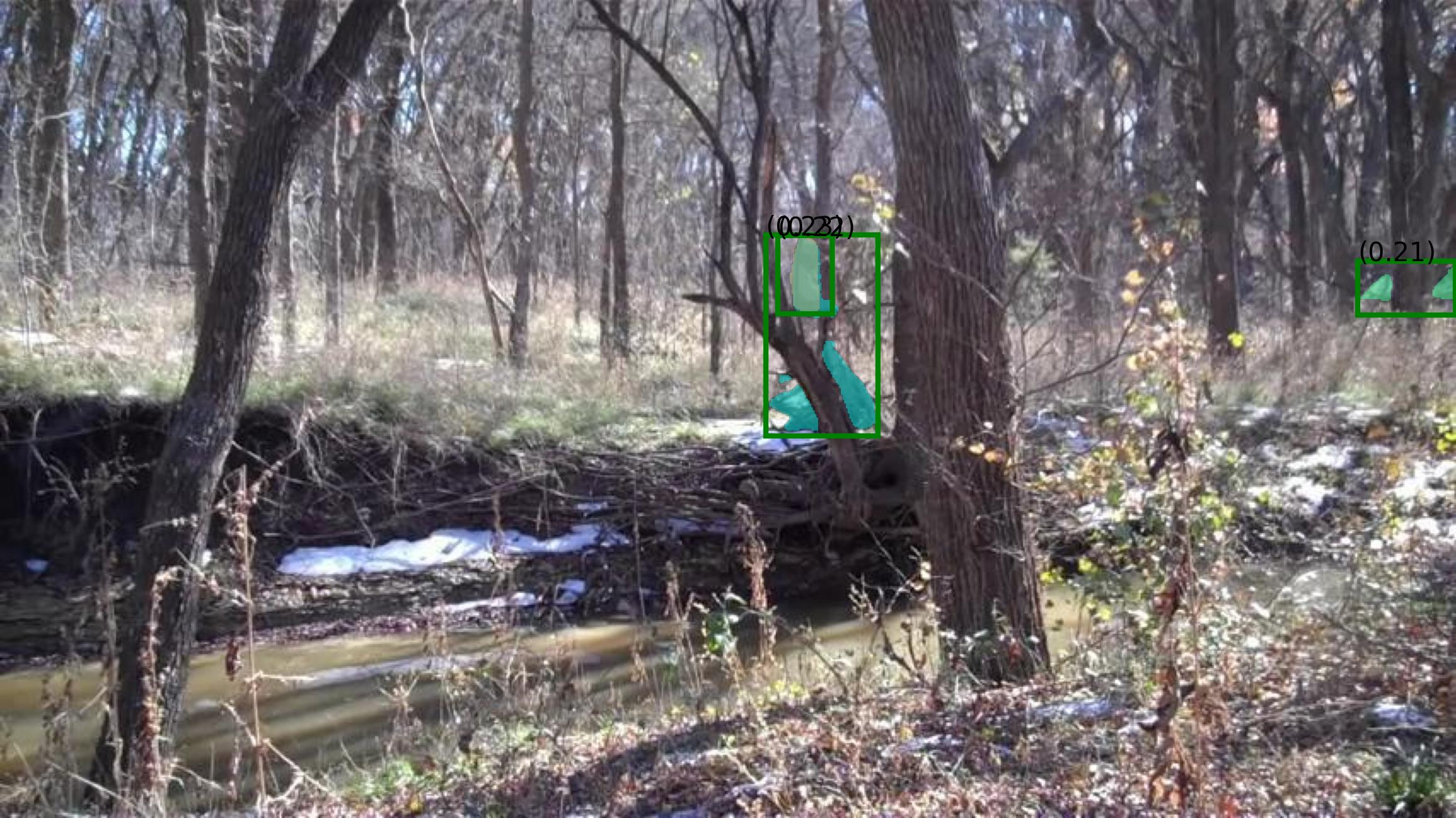}
    \end{subfigure}
    \begin{subfigure}{.16\textwidth}
        \centering
        \includegraphics[width=\textwidth]{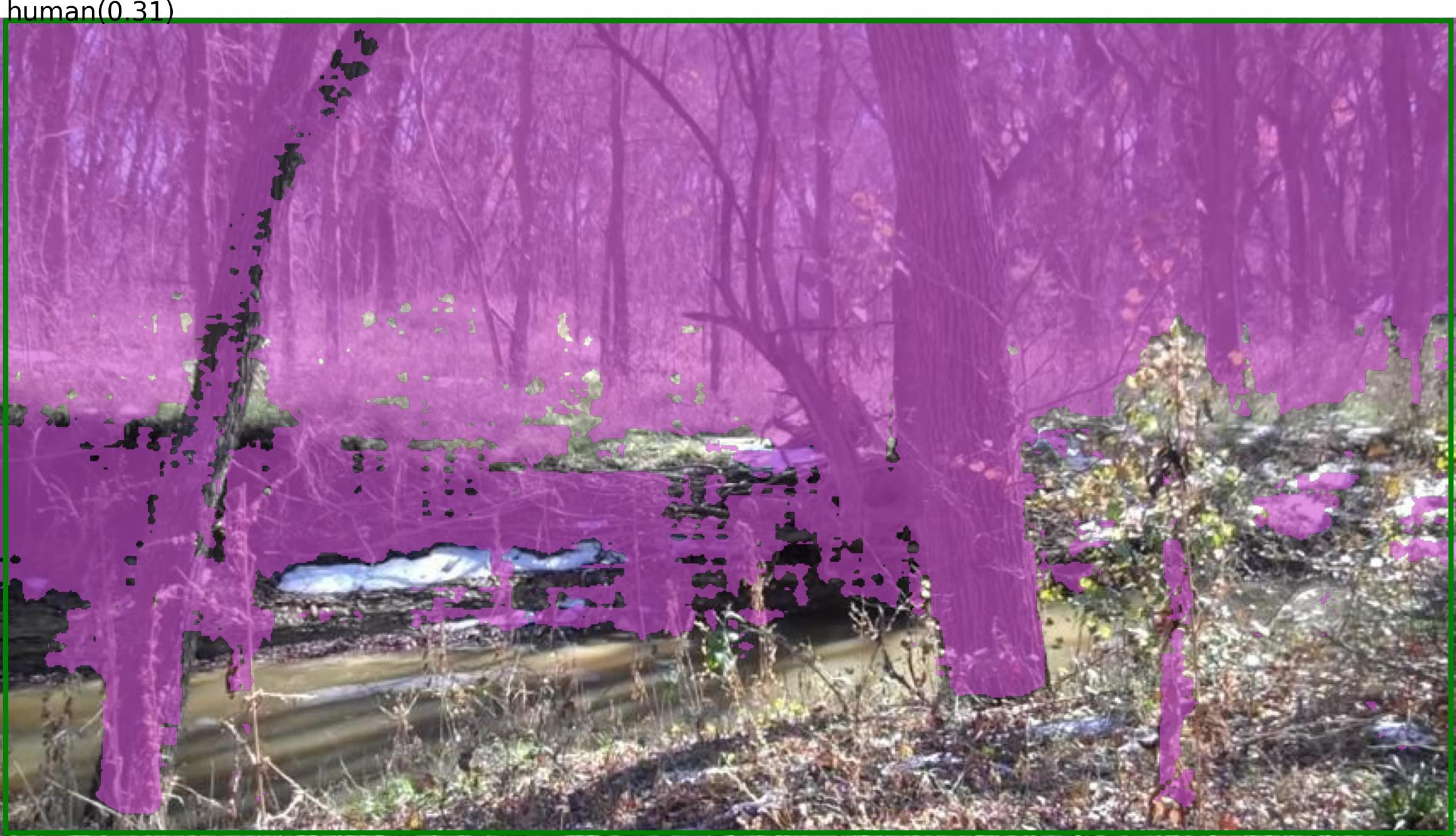}
    \end{subfigure}
    \begin{subfigure}{.16\textwidth}
        \centering
        \includegraphics[width=\textwidth]{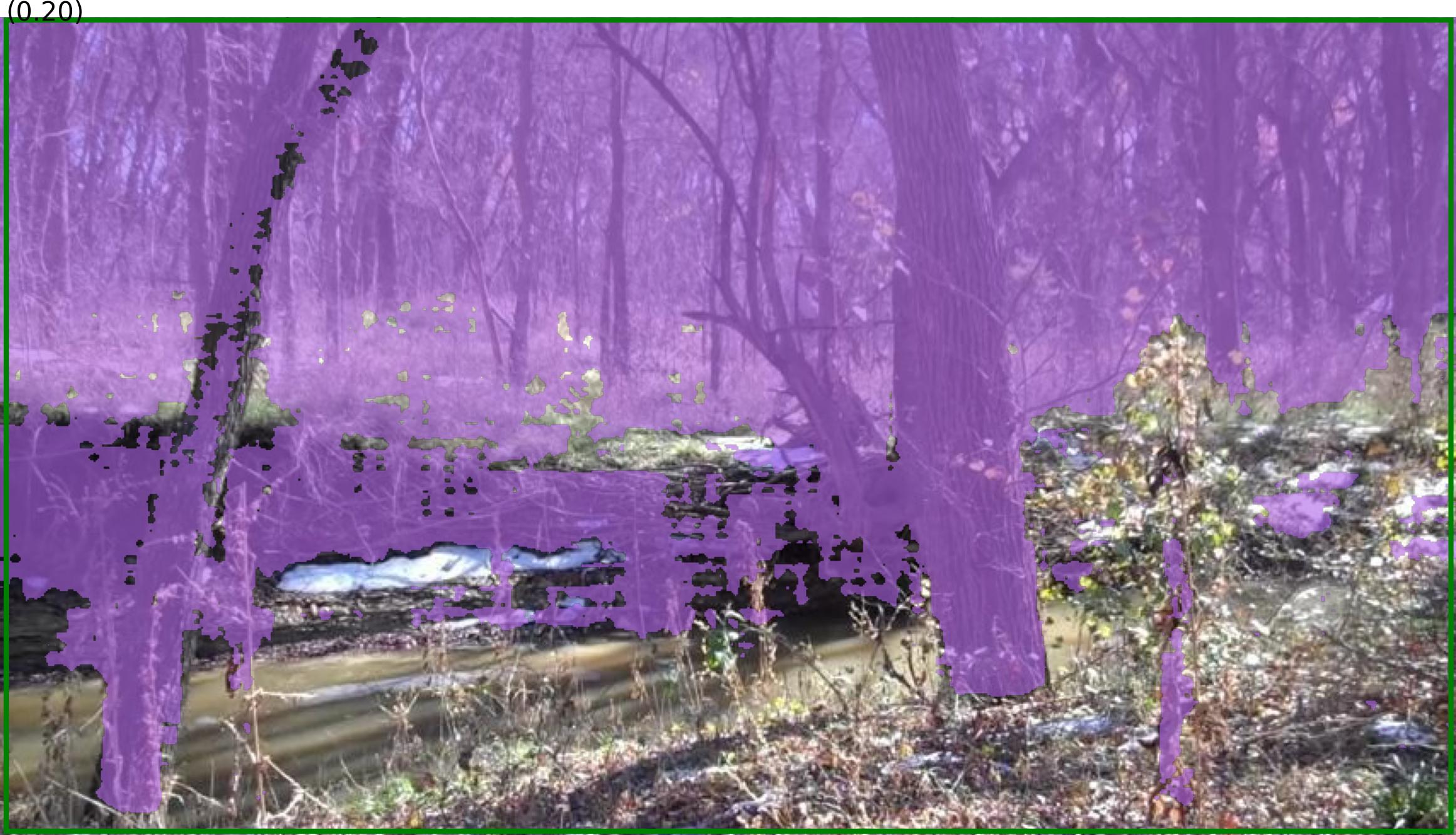}
    \end{subfigure}
    \begin{subfigure}{.16\textwidth}
        \centering
        \includegraphics[width=\textwidth]{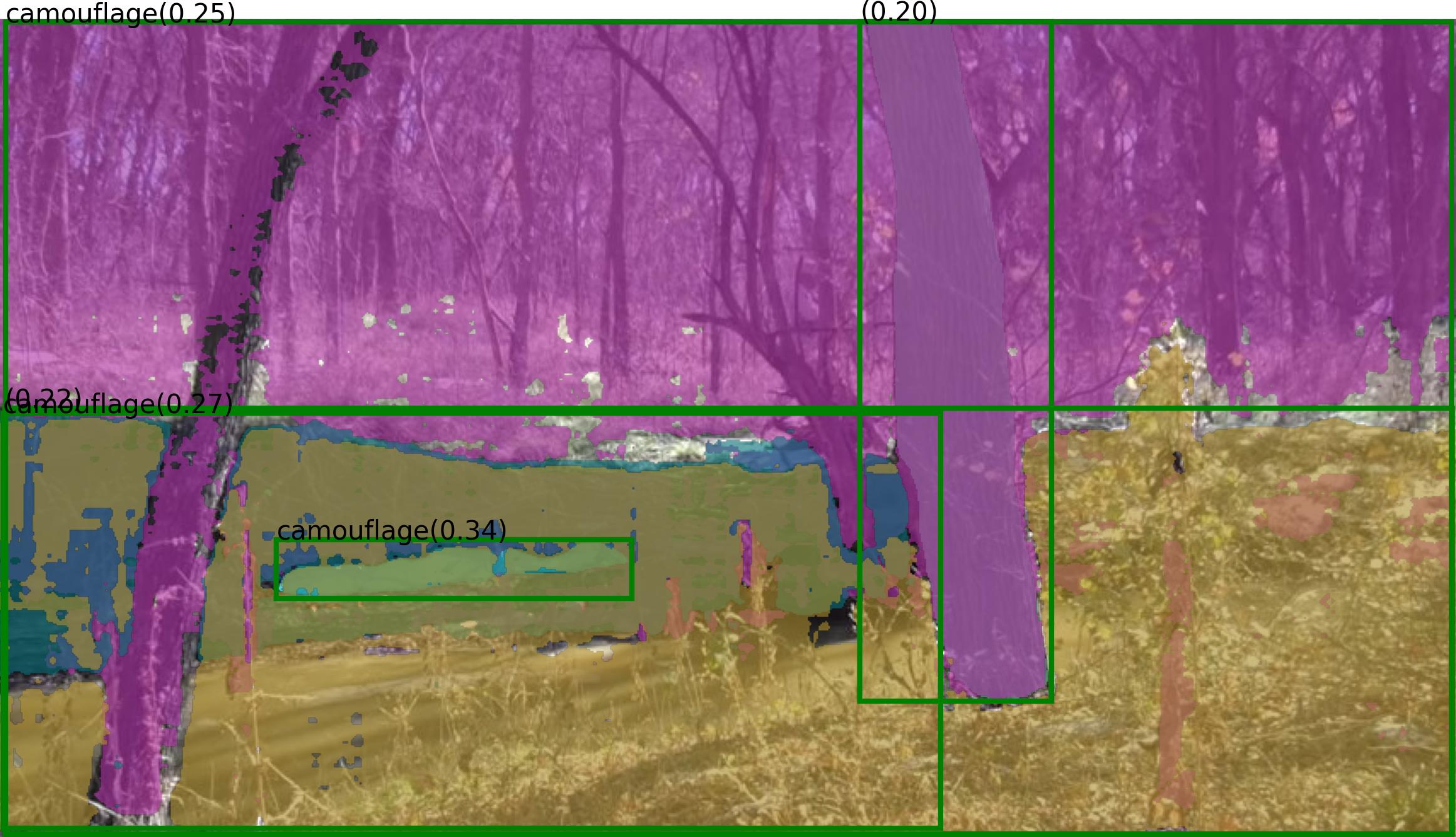}
    \end{subfigure}
    \begin{subfigure}{.16\textwidth}
        \centering
        \includegraphics[width=\textwidth]{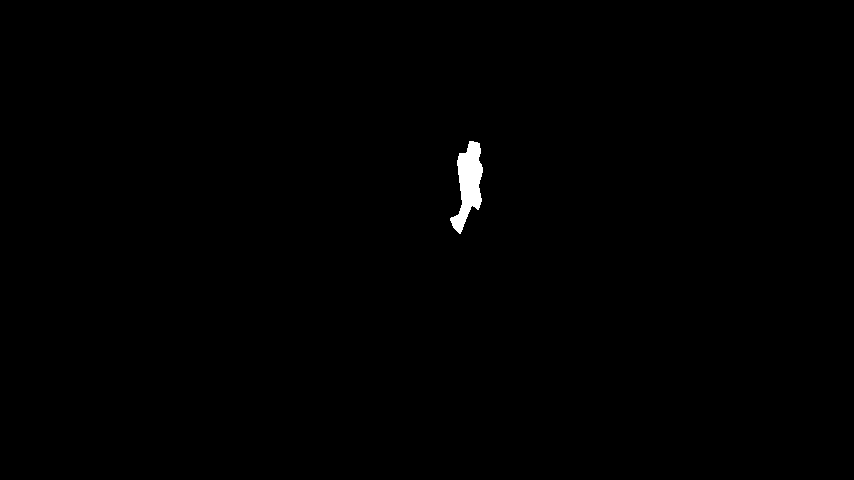}
    \end{subfigure}

    \begin{subfigure}{.16\textwidth}
        \centering
        \includegraphics[width=\textwidth]{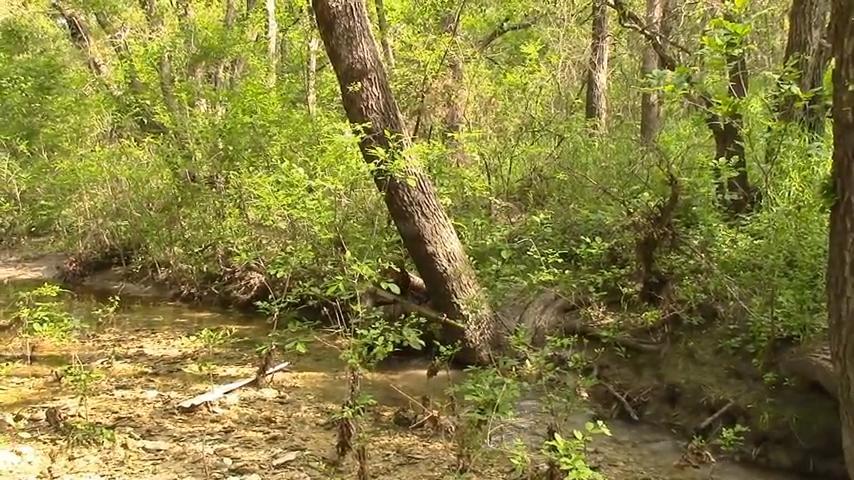}
        \caption{'Person'}
    \end{subfigure}
    \begin{subfigure}{.16\textwidth}
        \centering
        \includegraphics[width=\textwidth]{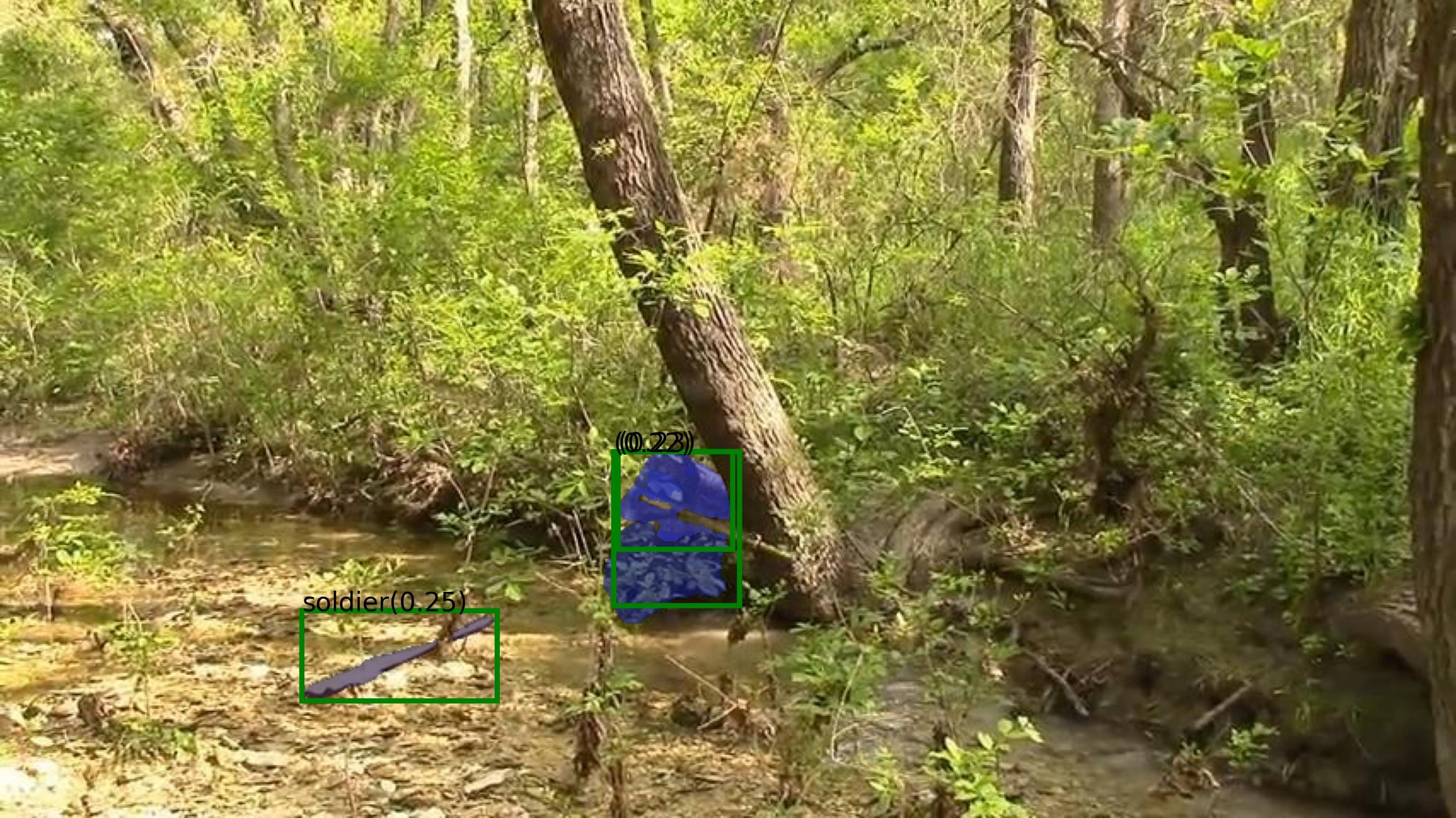}
        \caption{'Soldier'}
    \end{subfigure}
    \begin{subfigure}{.16\textwidth}
        \centering
        \includegraphics[width=\textwidth]{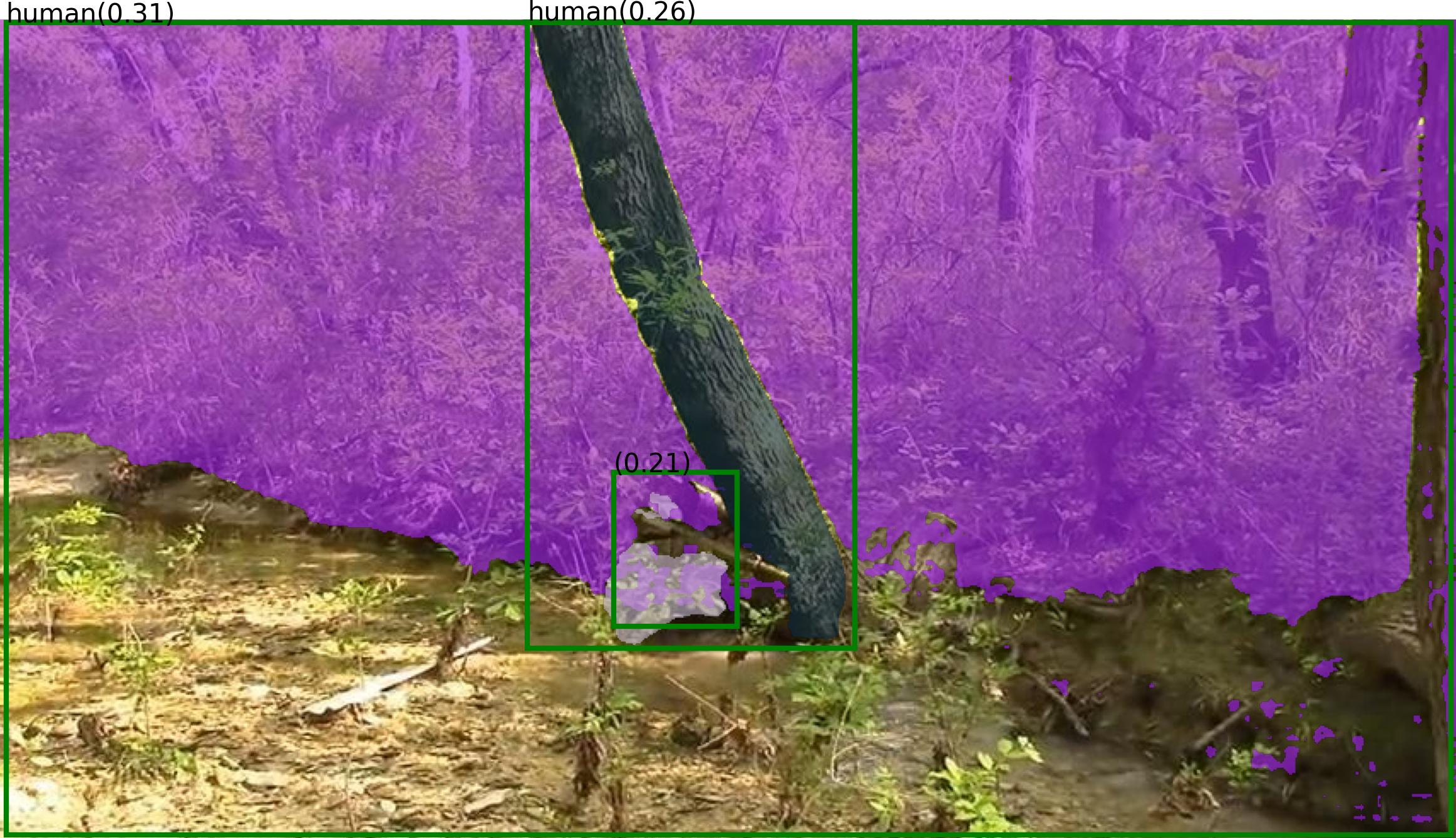}
        \caption{'Human'}
    \end{subfigure}
    \begin{subfigure}{.16\textwidth}
        \centering
        \includegraphics[width=\textwidth]{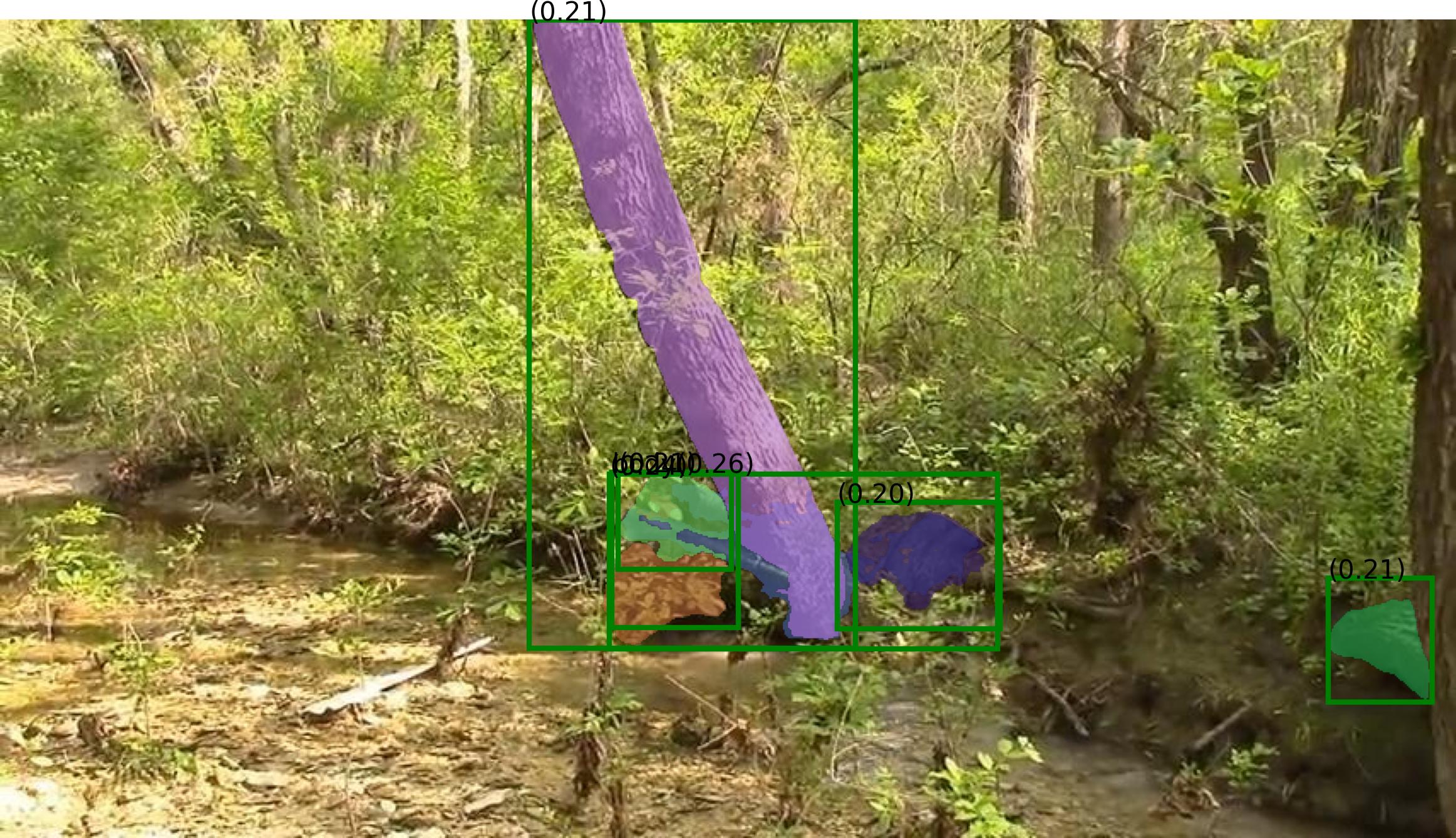}
        \caption{'Body'}
    \end{subfigure}
    \begin{subfigure}{.16\textwidth}
        \centering
        \includegraphics[width=\textwidth]{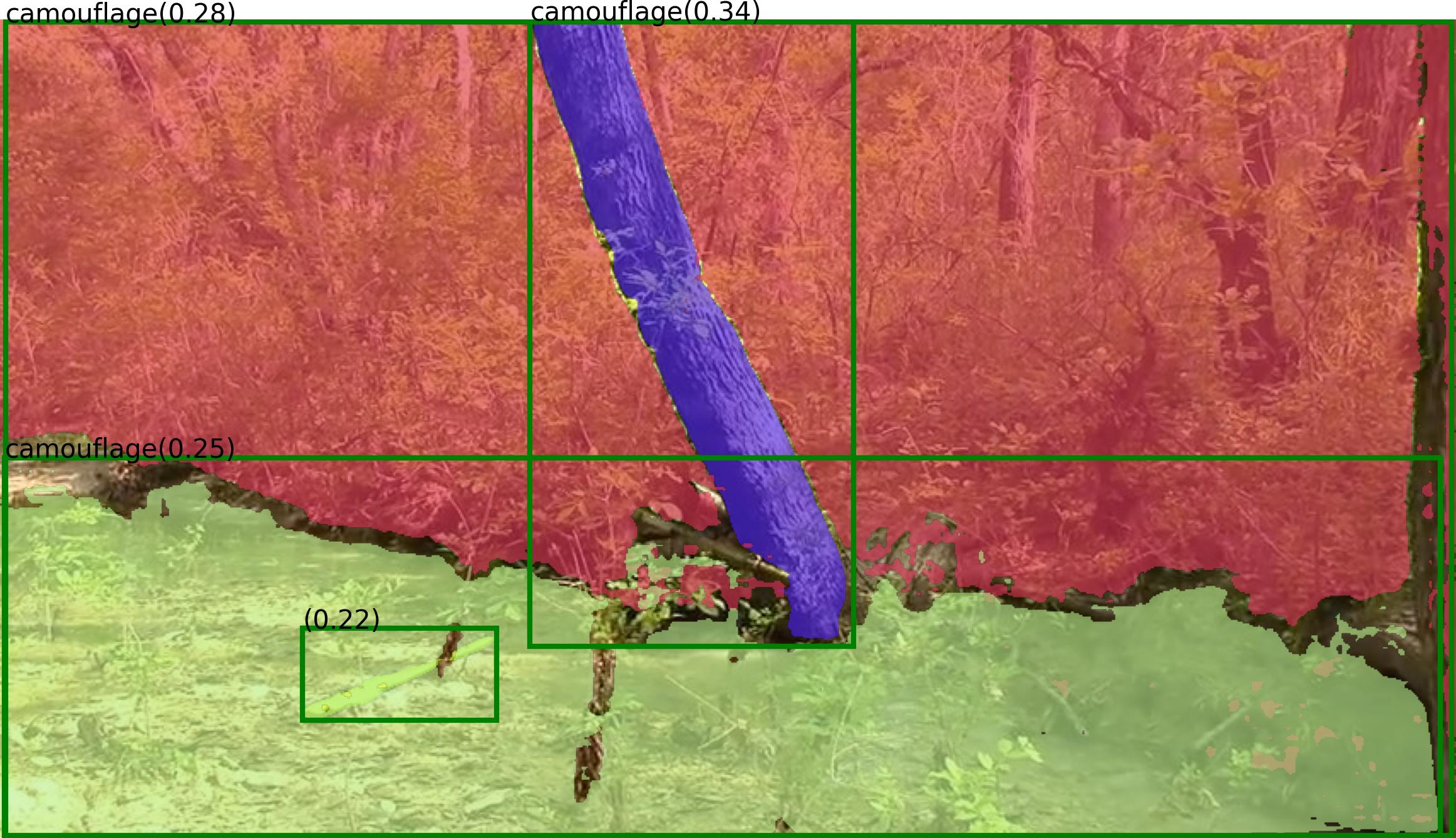}
        \caption{'Camouflage'}
    \end{subfigure}
    \begin{subfigure}{.16\textwidth}
        \centering
        \includegraphics[width=\textwidth]{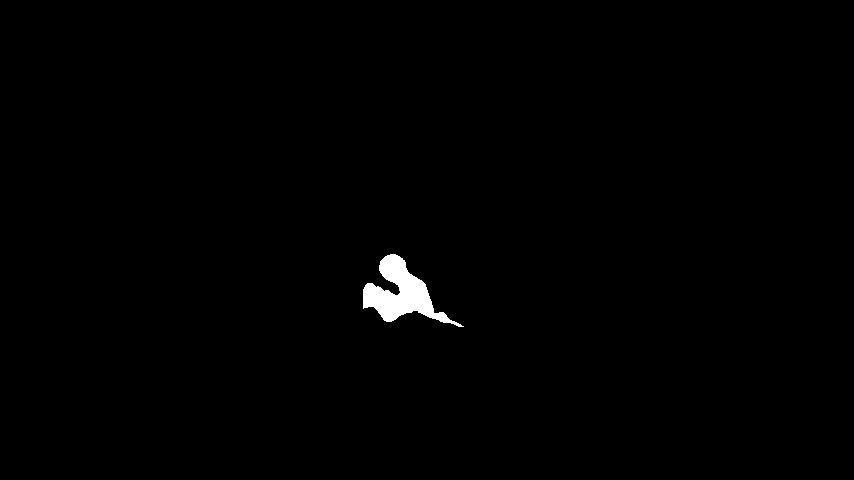}
        \caption{Ground truth}
    \end{subfigure}
    \caption{Different examples for the results of different phrases when prompting GSAM. Since camouflaged humans are difficult to spot, the only phrase that provides sufficient segmentation results is 'soldier'.}
    \label{fig:gsam_example_phrases}
\end{figure}

\noindent\textbf{GSAM:} since foundation models such as SAM are trained on a broad range of data and exhibit strong zero-shot performance on novel data~\cite{10388320}, we explore their zero-shot performance for pseudo-labeling in the challenging COD task. For our experiments we use the Grounding DINO weights provided by the authors and the ViT-L SAM model. Vision-language models often can be prompted with a text phrase to set the context~\cite{gu2023survey}. The phrase used for Grounding DINO has a big impact on the proposed bounding boxes. From a short qualitative comparison as shown in Fig.~\ref{fig:gsam_example_phrases}, we identify that the prompt 'soldier' produces good results. We also choose a low confidence threshold (0.15) for filtering the bounding boxes. This leads to at least one bounding box proposal per image, which is desired since CPD1K always contains one object instance per image. SAM is then used to pixel-precisely segment the object inside the bounding box. The resulting segmentation masks are then used as pseudo-labels for fine-tuning our models.

\noindent\textbf{Image inpainting using MAT:} image inpainting is used to fill (large) holes in an image with realistic content~\cite{Li.3292022}. We assume that if a mask is used, which covers the entire camouflaged object, the image inpainting model should use the appearance of the surrounding background to fill the camouflaged object with background content. By calculating the error between original and inpainted image, we should see a larger reconstruction error, where the camouflaged object was located in the original image. In the experiment, we use a sliding window approach with different window sizes to mask out and inpaint regions of the image. We use Mean Absolute Error (MAE) and Structural Similarity (SSIM)~\cite{wang2004ssim} to quantify the reconstruction error. Since MAT was trained on the Places365 dataset, which contains a large variety of different scenes, it should be able to fill in the holes in the forest environment of the CPD1K dataset with realistic content. Since MAT can only handle images at a resolution of $512\times512$\,pixels, we first downscale the images before further processing.

\noindent\textbf{HitNet:} HitNet is a state-of-the-art COD model that was trained on the COD10K and CAMO datasets. From its strong performance for COD on these datasets that contain camouflaged animals, it can be expected that the model generalizes well enough to also detect camouflaged humans, providing sufficiently good pseudo-labels.

\subsection{Learning from noisy labels}
\label{subsec:learning_with_noisy_labels}
Since pseudo-labels can vary strongly in their quality we propose and describe methods for either rejecting or down-weighting potentially weak pseudo-labels during training. We consider such low-quality pseudo-labels as either incorrectly localized and thus not overlapping with the real object or as overlapping with the real object but also segmenting background regions substantially larger than the object itself.

\noindent\textbf{Sample rejection:} we prevent the model from learning from bad pseudo-labels by separating the pseudo-labels in two sets $S_1, S_2$ with $S_1 \cap S_2 = \emptyset$ based on a well-suited quantitative label quality measure. $S_1$ will then be used for training while $S_2$ is discarded. The confidence value assigned to a bounding box by Grounding DINO is expected to be a promising quality metric. There are two potential approaches to reject potentially weak pseudo-labels: (1)~for each pseudo-label $\tilde Y$, we follow the assignment strategy as shown in Eq.~\ref{eq:conf_thresh}:
\begin{equation}
\begin{aligned}
    &\tilde Y \in S_1, \quad\text{if}\quad P(\tilde Y) \ge t_c\\
    &\tilde Y \in S_2, \quad\text{otherwise}
    \label{eq:conf_thresh}
\end{aligned}
\end{equation}
$t_c$ is the confidence threshold for sample rejection and all pseudo-labels with a confidence greater than this threshold are assigned to $S_1$, while the remaining pseudo-labels become element of $S_2$.\\(2)~A second potential quality measure can be the ratio of foreground and background pixels in an image according to the pseudo-label. We can assume to have exactly one camouflaged human per image. Since methods such as Ground SAM tend to segment large background regions as foreground in a failure case, we expect this to be a meaningful measure. The foreground ratio $ratio_{fg}$ can then be defined as shown in Eq.~\ref{eq:fg_ratio}:
\begin{equation}
    ratio_{fg}(\tilde Y) = \frac{1}{W \cdot H}\sum_{h = 1}^{H}\sum_{w = 1}^{W} \tilde y_{w,h}
    \label{eq:fg_ratio}
\end{equation}
$W$ and $H$ denote the width and height of the pseudo-label, respectively, and $\tilde y_{w,h}$ the pixel value of the binary segmentation mask at position $(w,h)$. This measure can then be used for thresholding to reject pseudo-labels that cover a larger foreground region than expected. This means that we assign all pseudo-labels with $ratio_{fg}(\tilde Y) \le t_f$ to $S_1$ and the remaining ones to $S_2$. $t_f$ is the threshold for the foreground ratio.

\noindent\textbf{Loss re-weighting:} inspired by Liu et al.~\cite{Liu_2016}, we reduce the influence of a bad pseudo-label in the training process. This is performed by utilizing the Grounding DINO confidences again as sufficiently well-suited quantitative label quality measure. This pseudo-label confidence, however, must be min-max scaled first as shown in Eq.~\ref{eq:min_max_scaling}:
\begin{equation}
    P(\tilde Y)' = \frac{P(\tilde Y) - \min(P(\tilde Y))}{\max(P(\tilde Y)) - \min(P(\tilde Y))}
    \label{eq:min_max_scaling}
\end{equation}
$P(\tilde Y)$ are the pseudo-label confidences assigned by Grounding DINO. The loss value $\mathcal{L}$ can then be scaled by the pseudo-label confidence $P(\tilde Y)'$ as shown in Eq.~\ref{eq:loss_reweighting}:
\begin{equation}
    \mathcal{L}' = \mathcal{L} \cdot P(\tilde Y)'
    \label{eq:loss_reweighting}
\end{equation}
This results in the most confident pseudo-label having the highest influence on the training process. Lower confidence pseudo-labels have less influence and the least confident pseudo-label is scaled to zero and thus it has no impact on the training.


\section{EXPERIMENTS AND RESULTS}
\label{sec:experiments}

The experiments are arranged in three parts: first, we set the baseline for fully supervised frugal learning-based COD. Then, we evaluate self-supervised learning to fine-tune pre-trained methods for camouflaged animal detection to the task of camouflaged human detection. Finally, we verify that pseudo-labels alone cannot solve COD and that self-supervised frugal fine-tuning is still necessary. We use the rarely used CPD1K dataset~\cite{Zheng.2019}, which contains images of camouflaged humans in different forest environments. Furthermore, we use two reference COD algorithms~\cite{Hu.3222022,Fan2020} that are pre-trained for camouflaged animal detection with the popular COD10K~\cite{Fan.b} and CAMO~\cite{Le.2019} datasets. For the quantitative evaluation, we calculate standard metrics for COD, namely the MAE, the $S$ measure~\cite{Fan.822017}, the $E_{\phi}$ measure~\cite{Fan.} and the $F_{\beta}^{w}$ measure~\cite{Margolin.}. We consider the MAE and the $F_{\beta}^{w}$ measure to be the most relevant for our experiments: MAE considers the True Negative (TN) pixels and thus reveals a segmentation approach that incorrectly segments the entire image as foreground. The $F_{\beta}^{w}$ measure does not consider TNs and thus provides a more balanced performance indication since the evaluation would be strongly biased towards TNs otherwise. Finally, there is a correlation between the $S$, the $E_{\phi}$, and the $F_{\beta}^{w}$ measure. To follow the standard COD evaluation protocols~\cite{Fan.b}, we consider all four measures in our summarizing table at the end of the section.

\subsection{Fully supervised frugal learning}
\label{subsec:frugalexperiments}
In this section, we work towards answering the question, which fraction of the CPD1K dataset's real training data is needed to fine-tune a pre-trained model from animal to human COD. It is desired to perform nearly on par with a model trained on the entire real training dataset. We evaluate the performance of two state-of-the-art COD models, namely HitNet~\cite{Hu.3222022} and SINet-V2~\cite{Fan.b}, under a frugal input data setup following the methodology explained in Section~\ref{subsec:camouflaged_frugal_learning}. While HitNet is transformer-based, SINet-V2 is CNN-based. In this way, we can compare, which of the two is better suited for frugal learning, considering the general assumption of transformer-based models requiring more data than CNN-based models to produce satisfying results.

\noindent\textbf{Experimental setup:} We randomly pick $k$ samples from the entire CPD1K training dataset to fine-tune the models. While the entire training set consists of 600 samples, we will not exceed $k=50$ meaning that we take about 10\,\% of the available training data in maximum. Due to this rather small sample sizes $k$, we run each experiment multiple times to account for random effects and to better understand the variance of the results. The cumulative mean and confidence interval values are calculated and tracked to verify this statistical robustness of our evaluation. We evaluate the models on the CPD1K test dataset. HitNet and SINet-V2 with pre-trained checkpoints provided by the authors are used as base models. Those are pre-trained on the combined training sets of COD10K and CAMO. Fine-tuning is done by using the same hyperparameter settings as the authors of HitNet and SINet-V2 were using in their corresponding papers. We only adapt the batch size if our $k$ is smaller than the original batch size value. We use the same training strategy and the same evaluation protocol used by the authors. After every epoch, validation is done on the CPD1K val dataset. Like the authors, we only keep the best training epoch and this checkpoint is then used for testing.

\begin{table}[ht]
    \centering
    \begin{tabular}{l|c|ccccccc|c}
        Method & Base model & $k=1$ & $k=2$ & $k=3$ & $k=5$ & $k=10$ & $k=30$ & $k=50$ & Fully fine-tuned \\
       \hline
       HitNet~\cite{Hu.3222022} & 0.564 & 0.567 & 0.584 & 0.591 & 0.618 & 0.673 & 0.734 & 0.753 & 0.828 \\
       SINet-V2~\cite{Fan.b} & 0.529 & 0.520 & 0.518 & 0.521 & 0.522 & 0.549 & 0.620 & 0.646 & 0.767\\
    \end{tabular}
    \caption{$k$-shot learning results for HitNet and SINet-V2 evaluated on the CPD1K test set with the $F_{\beta}^{w} \uparrow$ measure. HitNet generally outperforms SINet-V2 on this task. The base model indicates zero-shot performance.}
    \label{tab:kshot_results}
\end{table}
\noindent\textbf{Results:} The results are listed in Table~\ref{tab:kshot_results}. Besides different values for $k$, we also provide results for the \emph{base model}, which refers to applying the models in a zero-shot transfer setting, and for fine-tuning using the entire training dataset called \emph{fully fine-tuned}. For $k \le 5$ only little improvement is observable compared to the baseline. Starting from 10 shots there is steady improvement. When using 50 shots, the performance is already close to the fully fine-tuned model while only using about 10\,\% of the training data. For HitNet, the fully fine-tuned model improves the $F_{\beta}^{w}$ Measure by 46\,\%, the $k = 50$ model improves it by 33\,\%. There is also a big jump in performance between $k = 10$ and $k = 30$, which improves the performance by 9\,\%. When adding 20 more samples the performance improves by only 2.5\,\%. For SINet-V2 the fully fine-tuned model improves the baseline by 45\,\% while the model trained on 50 shots improves it by only 22\,\%. HitNet generally outperforms SINet-V2 on this task.

\begin{figure}[ht]
    \centering
    \includegraphics[width=\textwidth]{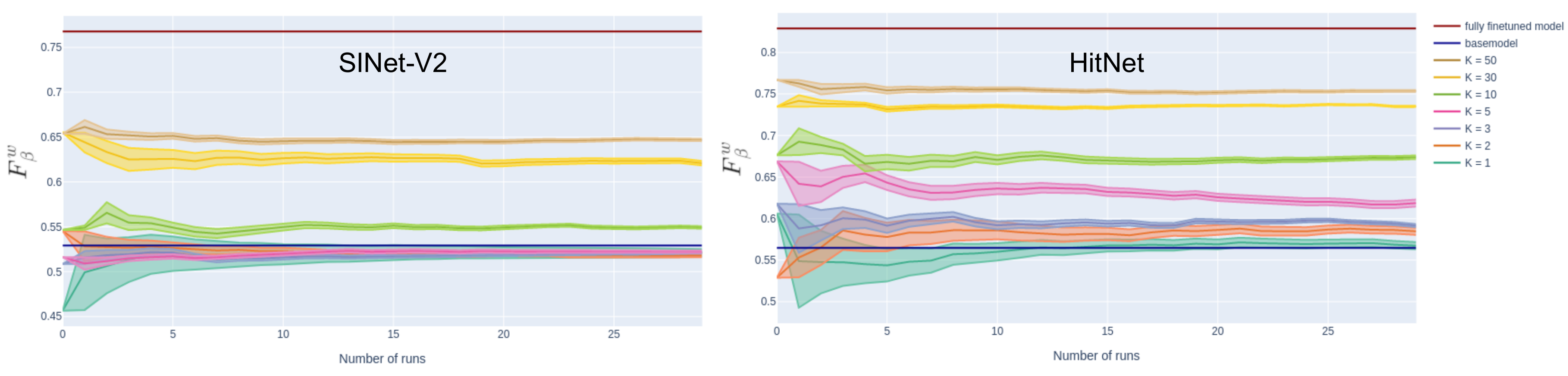}
    \caption{Cumulative mean $F_{\beta}^{w}\uparrow$ measure with 95\,\% confidence intervals across 30 repeated runs for SINet-V2 (left) and HitNet (right) and for $k=\{1,2,3,5,10,30,50\}$. After about 10 runs the mean value stabilizes. The performance ($F_{\beta}^{w}\uparrow$ measure) only slightly increases beyond $k=30$. HitNet outperforms SINet-V2. The relative gap between the fully fine-tuned and the frugally learned HitNet with $k=30$ is about 10\,\%.}
    \label{fig:fsod_wfm}
\end{figure}

Fig.~\ref{fig:fsod_wfm} shows the development of the $F_{\beta}^{w}$ measure over multiple training runs for different values of $k$. This means that we run the same experiment for the same value of $k$ multiple times to achieve statistical stability. As the performance of the trained model in this frugal learning setup strongly depends on the few samples randomly picked from the full training dataset, we can see a rather high variance in the $F_{\beta}^{w}$ measure during the first runs. For smaller values of $k$, this variance is obviously even higher. It becomes apparent that such an analysis is necessary to prevent an insufficient statistical analysis under scarce data settings such as frugal learning or FSOD~\cite{Wang.3162020}. After about 10 runs the mean value stabilizes. For both models there seems to be a sweet spot at $k=30$ providing good performance according to the $F_{\beta}^{w}$ measure while using only a fraction of the training data of about 6\,\%. The improvement gained by $k=50$ is rather small compared to $k=30$.

\begin{figure}[ht]
    \centering
    \includegraphics[width=\textwidth]{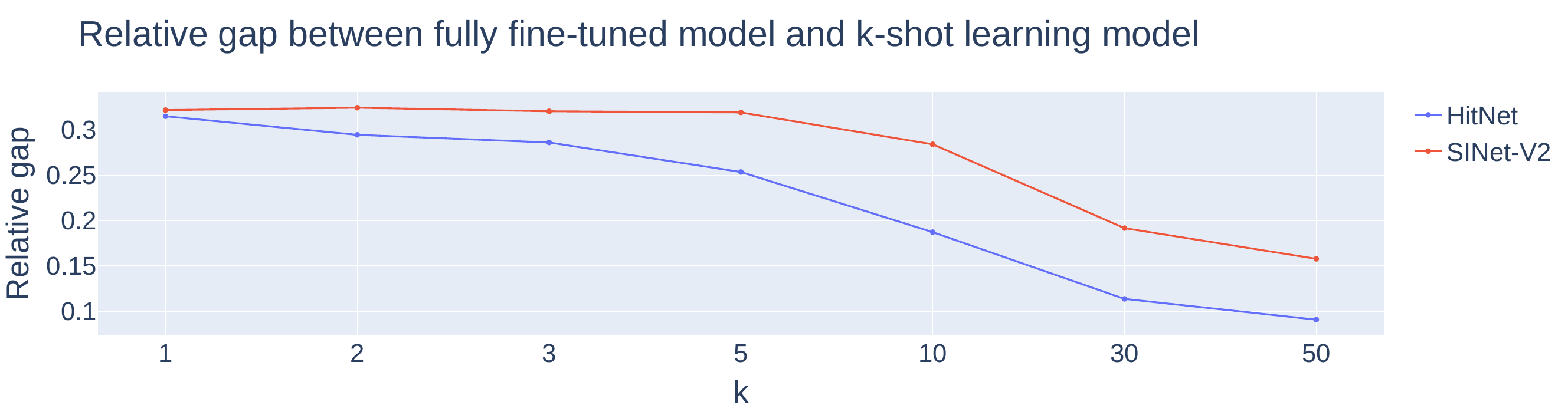}
    \caption{Relative gap between the fully fine-tuned models and the $k$-shot models for HitNet and SINet-V2 based on the $F_{\beta}^{w}$ measure. HitNet is able to narrow this gap clearly better compared to SINet-V2.}
    \label{fig:kshot_gap}
\end{figure}

Besides the absolute performance as shown in Fig.~\ref{fig:fsod_wfm}, we also compare the relative performance gap between the fully fine-tuned models and the $k$-shot models for HitNet and SINet-V2 based on the $F_{\beta}^{w}$ measure. The results are visualized in Fig.~\ref{fig:kshot_gap}. HitNet is able to narrow this gap clearly better compared to SINet-V2. The relative gap between the fully fine-tuned and the frugally learned HitNet with $k=30$ is about 10\,\%. We can summarize that HitNet provides better absolute and relative performance compared to SINet-V2. As a result, we will focus on HitNet frugally learned with $k=30$ in the subsequent experiments.


\subsection{Self-supervised learning from pseudo-labels}
\label{sec:experiments_ssl}

In this section, we will explore how we can utilize unlabeled data for the COD task. First, we will set a baseline using the three methods for generating pseudo-labeling described in Section~\ref{subsec:learning_with_noisy_labels}. Then, we explore different methods for improving this baseline. Those generated pseudo-labels are used to train HitNet and SINet-V2 and we quantitatively compare their performance in ablation studies using the GT labels.

\noindent\textbf{Baseline quality assessment:} We start with a qualitative evaluation of the three pseudo-labeling methods. They are first evaluated separately, before we compare them quantitatively in an ablation study.

\begin{figure}[ht]
\centering
    \begin{subfigure}{.325\textwidth}
        \centering
        \includegraphics[width=\textwidth]{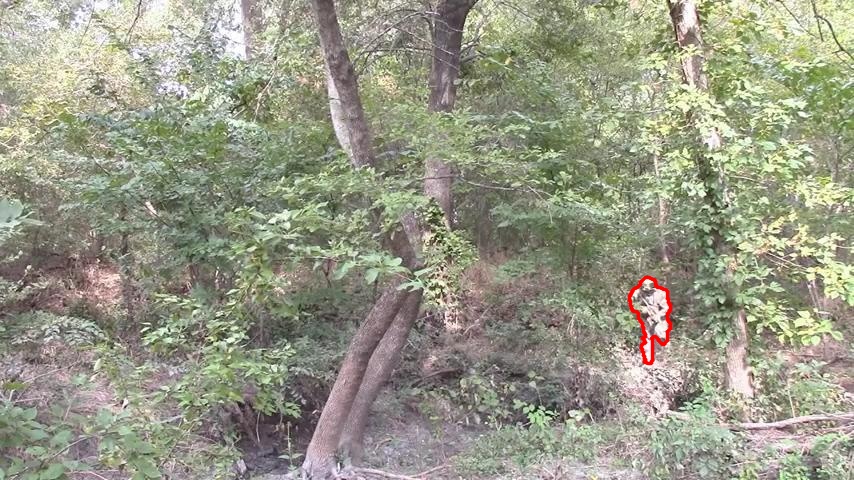}
    \end{subfigure}
    \begin{subfigure}{.325\textwidth}
        \centering
        \includegraphics[width=\textwidth]{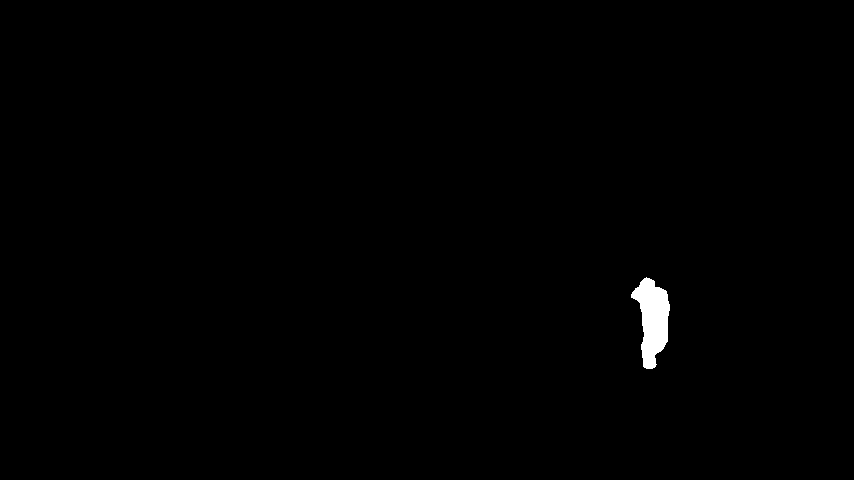}
    \end{subfigure}
    \begin{subfigure}{.325\textwidth}
        \centering
        \includegraphics[width=\textwidth]{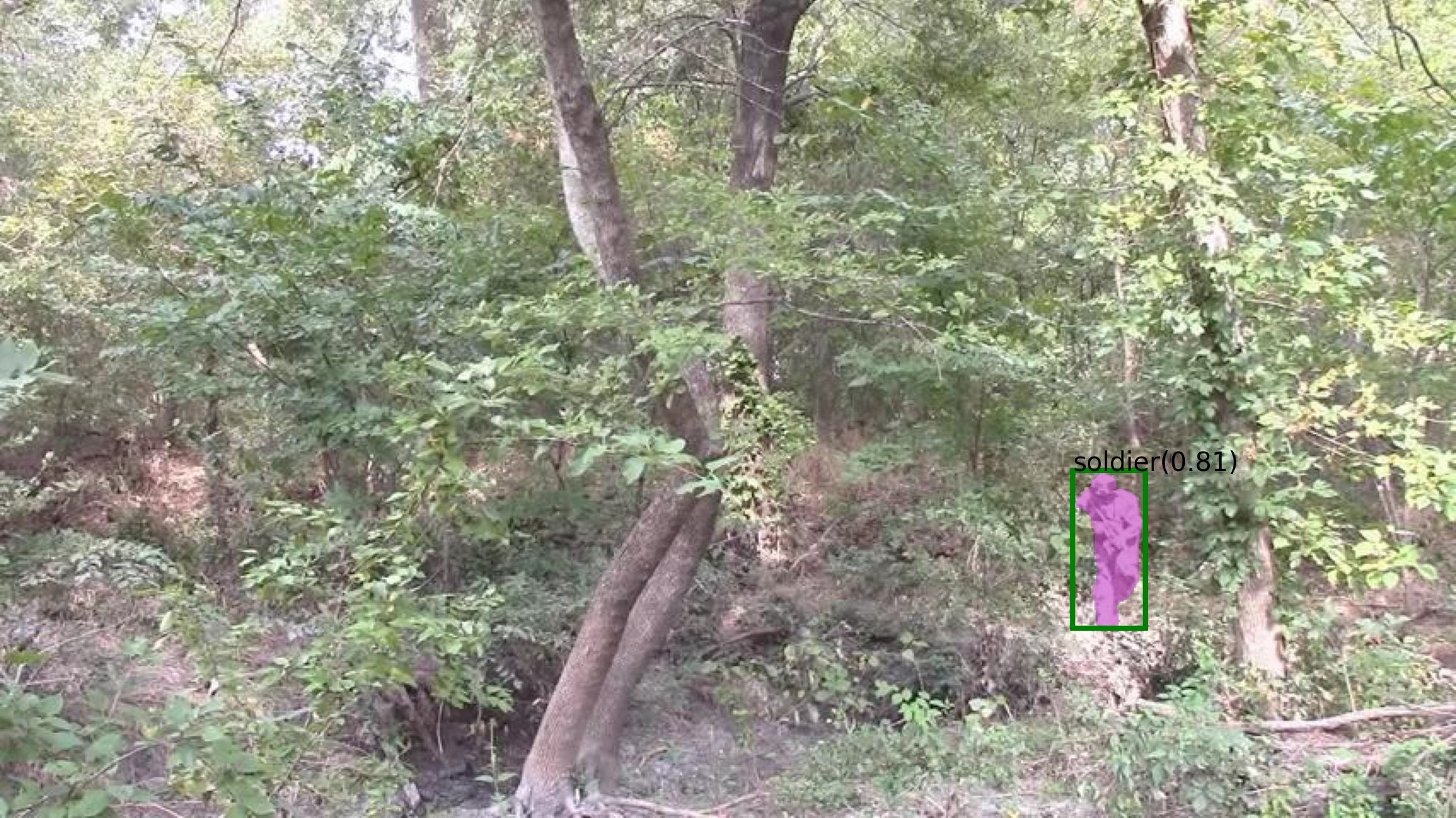}
    \end{subfigure}

    \begin{subfigure}{.325\textwidth}
        \centering
        \includegraphics[width=\textwidth]{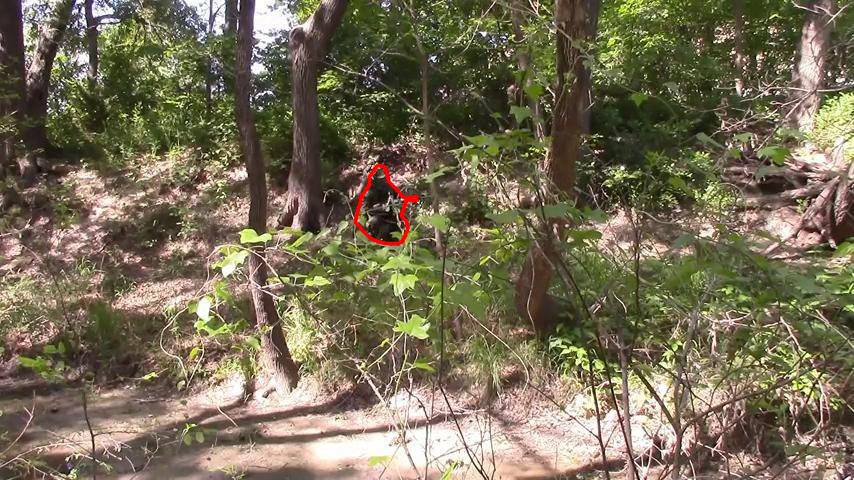}
        \caption{Input image}
    \end{subfigure}
    \begin{subfigure}{.325\textwidth}
        \centering
        \includegraphics[width=\textwidth]{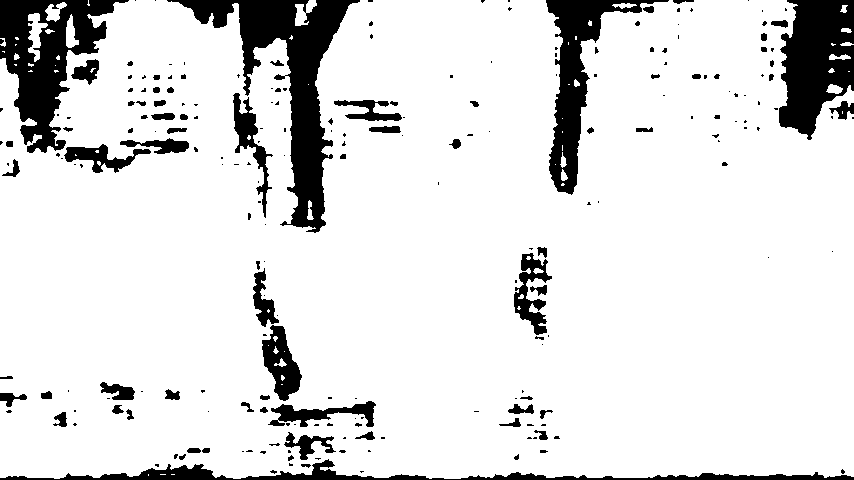}
        \caption{Segmentation map}
    \end{subfigure}
    \begin{subfigure}{.325\textwidth}
        \centering
        \includegraphics[width=\textwidth]{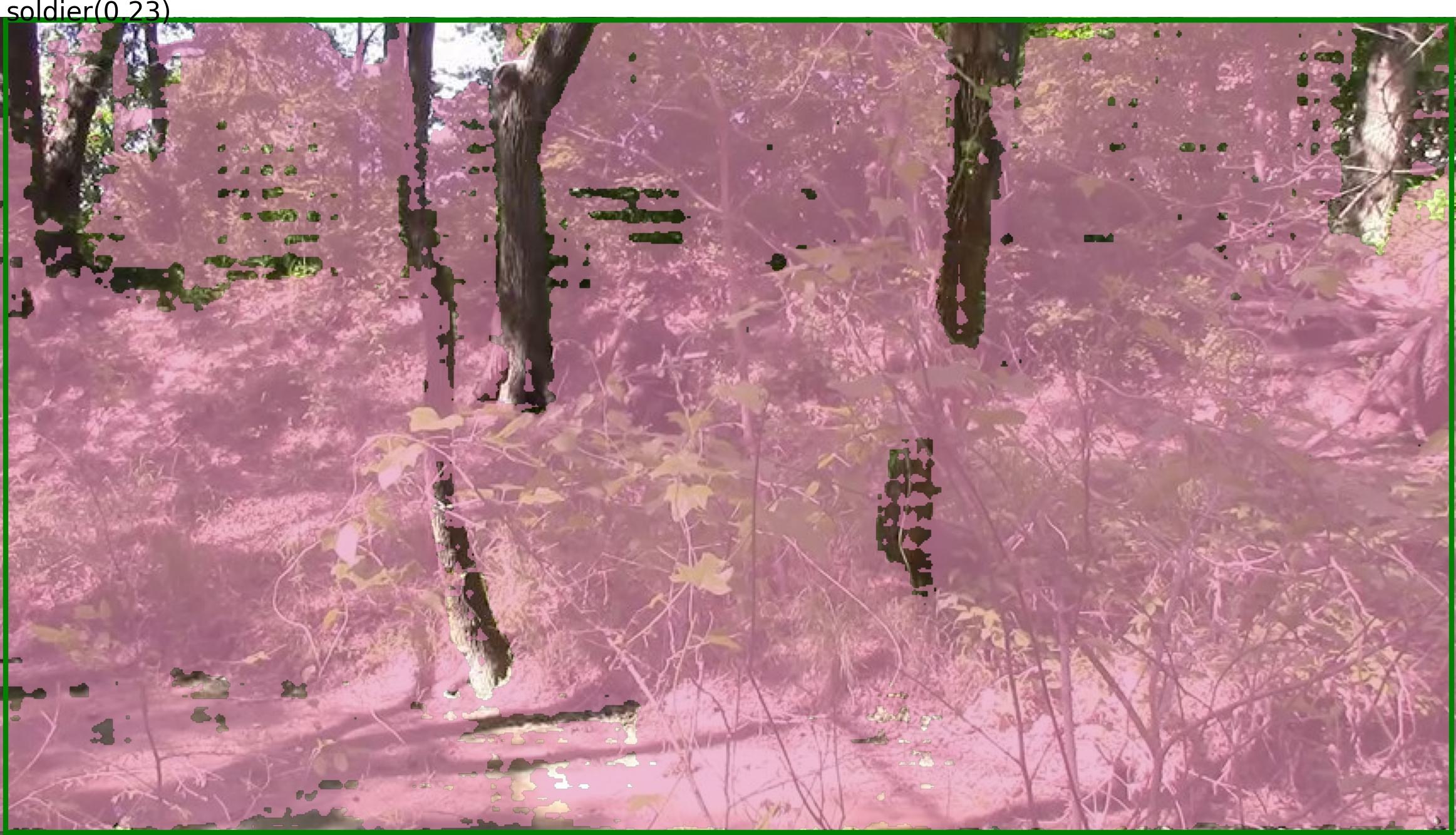}
        \caption{Bounding box proposal}
    \end{subfigure}
    \caption{Example output of GSAM prompted with the phrase 'soldier'. In the first row, the soldier is located well by Grounding DINO, which leads to SAM generating a mostly correct segmentation map that can serve well as a pseudo-label. In the second row, the soldier is not detected and the incorrect bounding box proposal leads to an arbitrary segmentation map. The human is indicated in red color in the input image.}
    \label{fig:sam_pseudolabels}
\end{figure}

\begin{figure}[ht]
    \centering
    \begin{subfigure}{.19\textwidth}
        \centering
        \includegraphics[width=\textwidth]{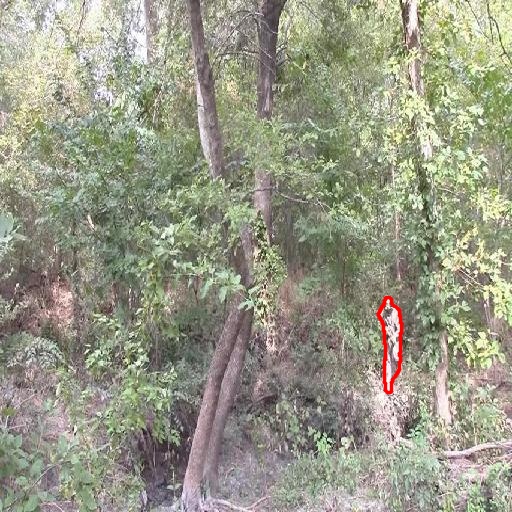}
    \end{subfigure}
    \begin{subfigure}{.19\textwidth}
        \centering
        \includegraphics[width=\textwidth]{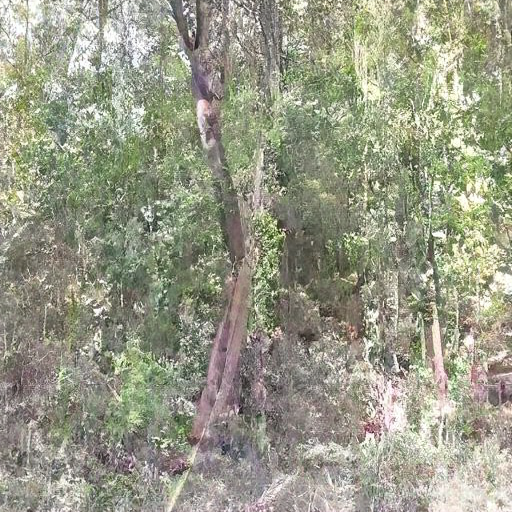}
    \end{subfigure}
    \begin{subfigure}{.19\textwidth}
        \centering
        \includegraphics[width=\textwidth]{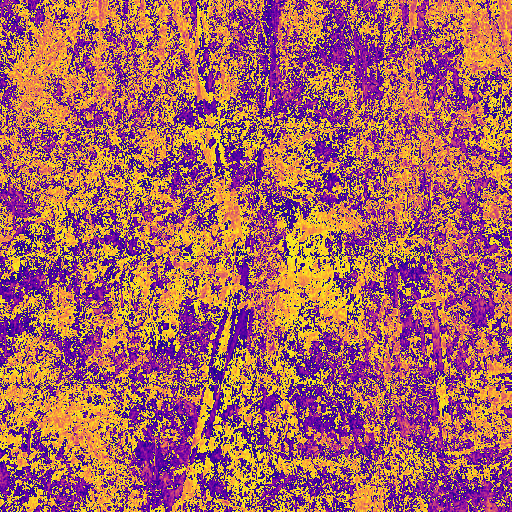}
    \end{subfigure}
    \begin{subfigure}{.19\textwidth}
        \centering
        \includegraphics[width=\textwidth]{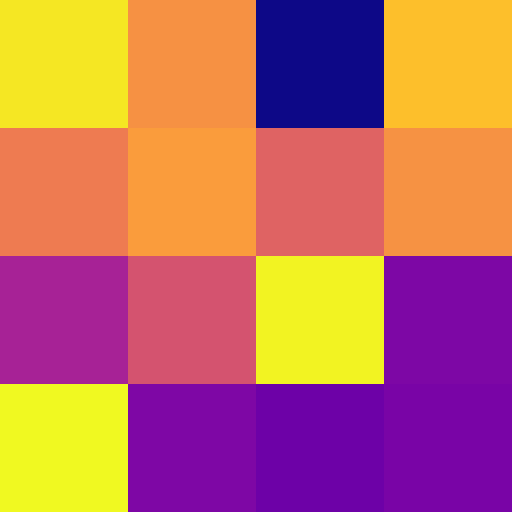}
    \end{subfigure}
    \begin{subfigure}{.19\textwidth}
        \centering
        \includegraphics[width=\textwidth]{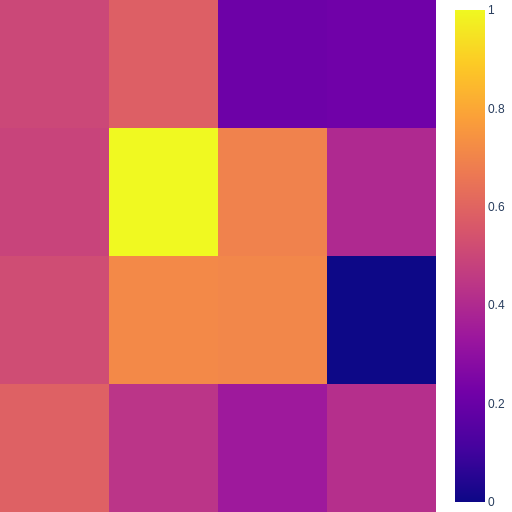}
    \end{subfigure}

    \begin{subfigure}{.19\textwidth}
        \centering
        \includegraphics[width=\textwidth]{images/ssl/mat_examples/dataset01_01_00002310_512512_contour.jpg}
    \end{subfigure}
    \begin{subfigure}{.19\textwidth}
        \centering
        \includegraphics[width=\textwidth]{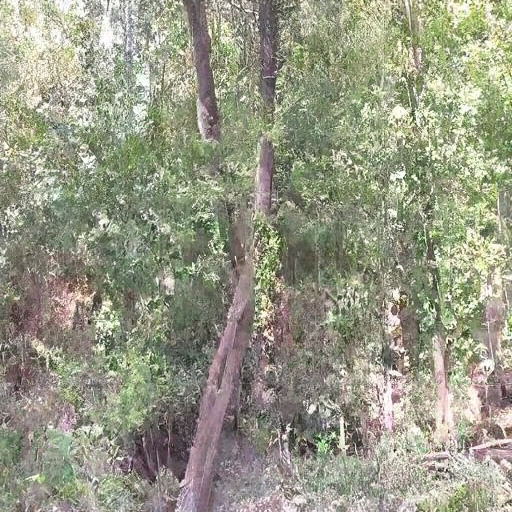}
    \end{subfigure}
    \begin{subfigure}{.19\textwidth}
        \centering
        \includegraphics[width=\textwidth]{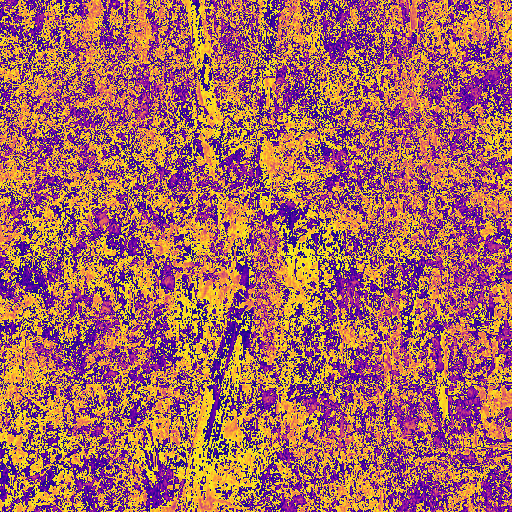}
    \end{subfigure}
    \begin{subfigure}{.19\textwidth}
        \centering
        \includegraphics[width=\textwidth]{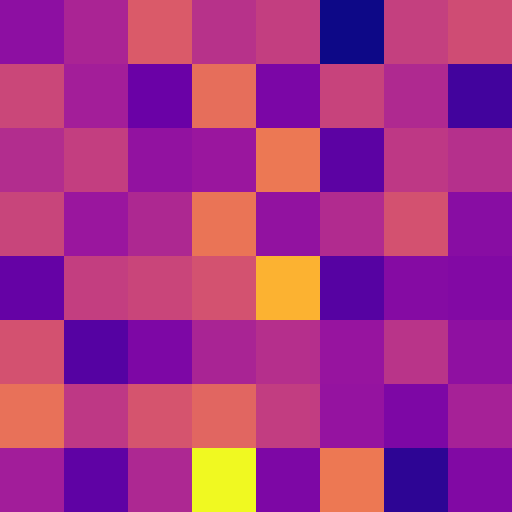}
    \end{subfigure}
    \begin{subfigure}{.19\textwidth}
        \centering
        \includegraphics[width=\textwidth]{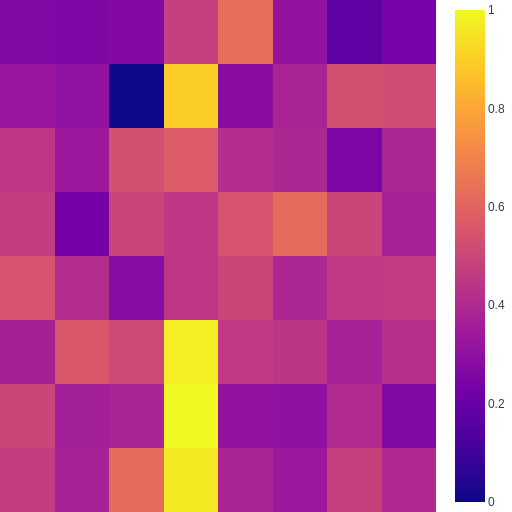}
    \end{subfigure}

    \begin{subfigure}{.19\textwidth}
        \centering
        \includegraphics[width=\textwidth]{images/ssl/mat_examples/dataset01_01_00002310_512512_contour.jpg}
        \caption{Input image}
    \end{subfigure}
    \begin{subfigure}{.19\textwidth}
        \centering
        \includegraphics[width=\textwidth]{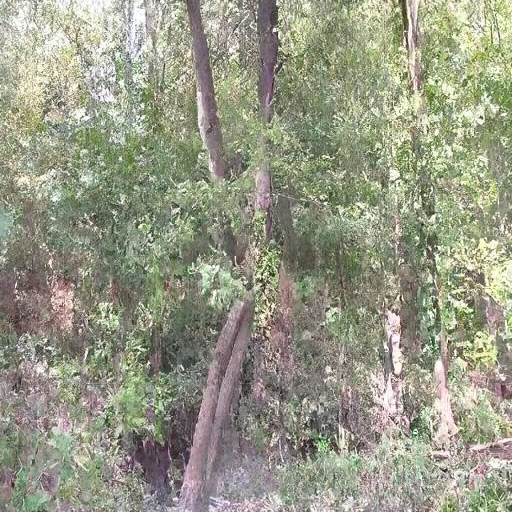}
        \caption{Inpainted image}
    \end{subfigure}
    \begin{subfigure}{.19\textwidth}
        \centering
        \includegraphics[width=\textwidth]{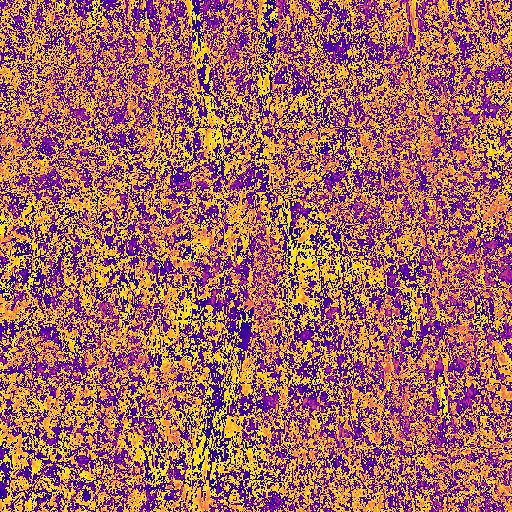}
        \caption{Pixel similarity}
    \end{subfigure}
    \begin{subfigure}{.19\textwidth}
        \centering
        \includegraphics[width=\textwidth]{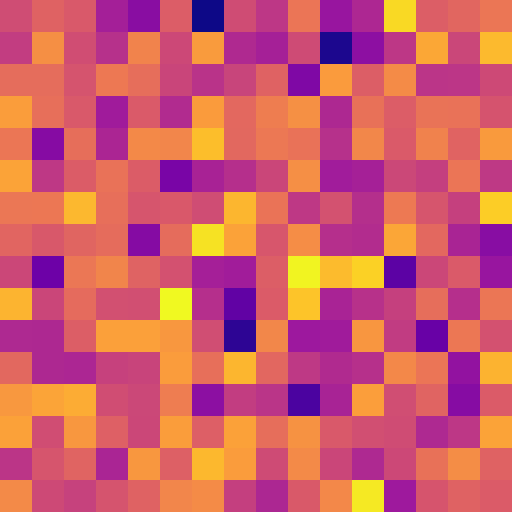}
        \caption{Region similarity}
    \end{subfigure}
    \begin{subfigure}{.19\textwidth}
        \centering
        \includegraphics[width=\textwidth]{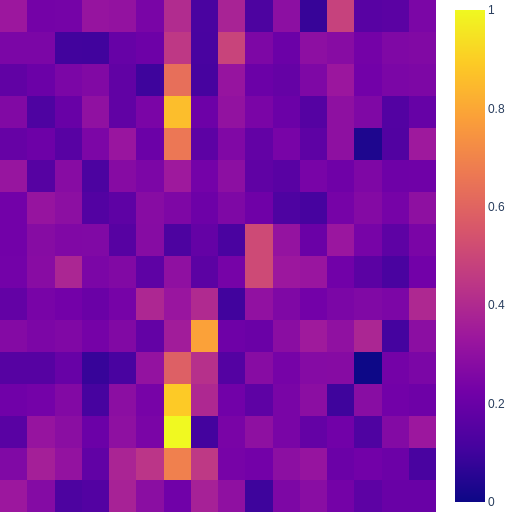}
        \caption{SSIM}
    \end{subfigure}

    \caption{Results of MAT-based image inpainting for different tile sizes: first row 128px, second row 64px, third row 32px. Three methods are tested to measure region similarities: (c) pixel-error based pixel similarity, (d) MAE-based region similarity, and (e) SSIM. The assumed low similarity would be indicated by a dark blue color. None of the tested approaches meets this assumption.}
    \label{fig:mat_pseudolabel}
\end{figure}

\begin{figure}[ht]
    \centering
    \begin{subfigure}{.48\textwidth}
        \centering
        \includegraphics[width=\textwidth]{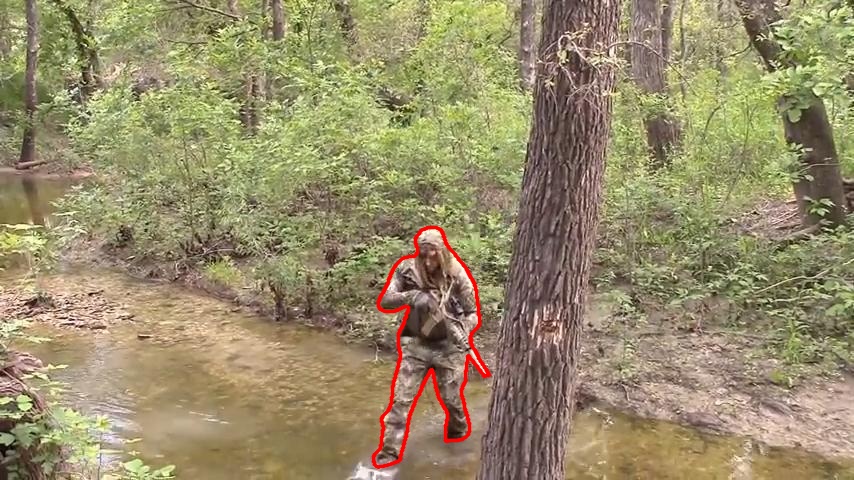}
    \end{subfigure}
    \begin{subfigure}{.48\textwidth}
        \centering
        \includegraphics[width=\textwidth]{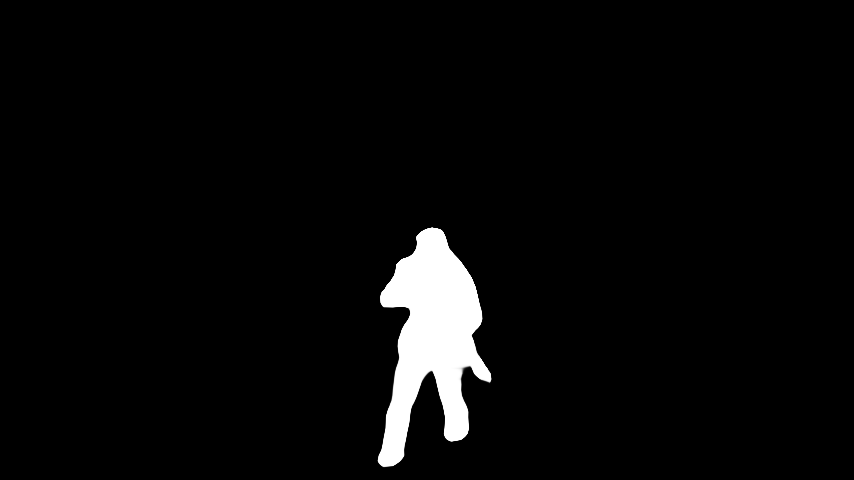}
    \end{subfigure}

    \begin{subfigure}{.48\textwidth}
        \centering
        \includegraphics[width=\textwidth]{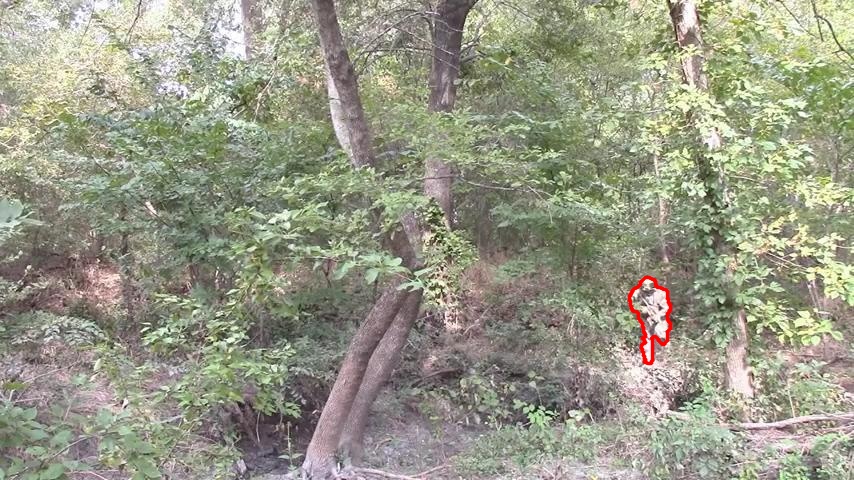}
        \caption{Input image}
    \end{subfigure}
    \begin{subfigure}{.48\textwidth}
        \centering
        \includegraphics[width=\textwidth]{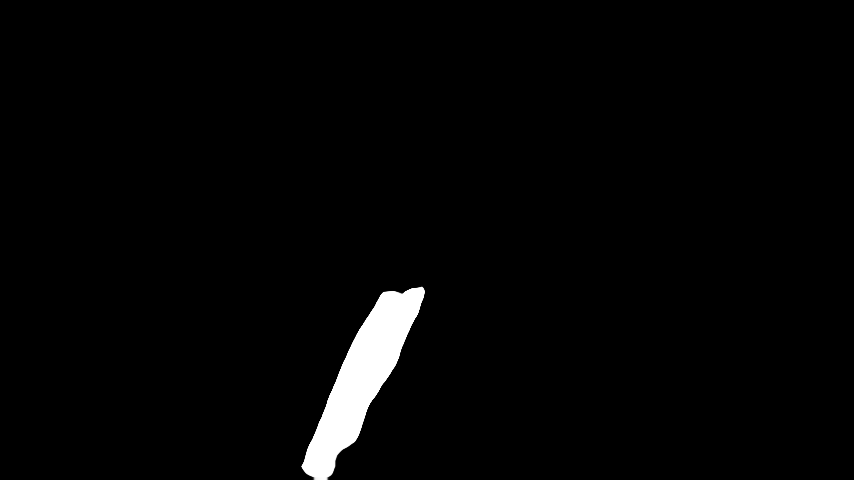}
        \caption{Segmentation map}
    \end{subfigure}
    \caption{Example output provided by the original HitNet pre-trained for camouflaged animal detection. In the first row, the human is segmented decently, while in the second row HitNet fails to segment the human. HitNet is not trained on camouflaged humans but on animals. We assume that the structure of the tree might look like a camouflaged animal sitting on bark. Thus, HitNet predicts it to be a camouflaged object.}
    \label{fig:hitnet_pseudolabel}
\end{figure}

\begin{itemize}
\item \textbf{GSAM:} Fig.\ref{fig:sam_pseudolabels} shows two examples for pseudo-labeling using GSAM~\cite{ren2024grounded}. In the first row the soldier is precisely located. This is also indicated by the high confidence of 0.81 that Grounding DINO assigned to the proposal. SAM is then able to use the proposal to accurately segment the camouflaged human. In the second row, Grounding DINO is not able to detect the human but it still provides a bounding box proposal. Since SAM is trained to generate at least one segmentation mask given an input, it generates a segmentation map mostly segmenting the background vegetation. There is potential in this approach, so we consider it for the quantitative evaluation.
\item \textbf{Image inpainting:} In Fig.~\ref{fig:mat_pseudolabel}, the performance of image inpainting using the MAT~\cite{Li.3292022} approach is visualized. The image is subdivided in tiles and each tile is then masked and inpainted using the remainder of the input image. The similarity between related tiles (i.e. same position) in the original and the inpainted image is calculated. We assume that the position of the camouflaged human in the image produces a measurable dissimilarity compared to the background regions. We test three measures: (1)~the pixel similarity which is the inverse of the pixel difference, (2)~the region similarity calculated by the inverse of the MAE, and (3)~the SSIM. The results in Fig.~\ref{fig:mat_pseudolabel} show that none of the proposed metrics can be used as a prior for the tile to contain an object. In contrast to our assumption that the similarity is low at the position of the human, the similarity is sometimes even higher compared to the surrounding regions as indicated by the bright yellow color in the first row's example. We can summarize that image inpainting cannot serve as a promising source of pseudo-labels.
\item \textbf{Baseline HitNet:} Here we use HitNet pre-trained on COD10K and CAMO for camouflaged animal detection to generate pseudo-labels for camouflaged human detection. Fig.~\ref{fig:hitnet_pseudolabel} shows that this approach works well if the human is sufficiently salient in the image (upper row). In a more complex scene (lower row), HitNet completely fails to segment the human. This is probably the case because HitNet is not trained on camouflaged humans but on animals. The structure of the tree then might resemble a camouflaged animal sitting on bark and HitNet predicts it to be a camouflaged object. We do see some potential in this approach. Thus, we consider it for the quantitative evaluation.
\end{itemize}

Extracting pseudo labels seems promising with the GSAM and the Baseline HitNet approach. Since image inpainting does not provide meaningful results, it is disregarded for the quantitative evaluation. Table~\ref{tab:pseudo_label_comp} shows this quantitative comparison of the two remaining approaches. GSAM outperforms Baseline HitNet by a large margin according to the $F_{\beta}^{w}$ measure. Besides the $F_{\beta}^{w}$ measure, however, we also use the MAE as evaluation measure. Interestingly, Baseline HitNet clearly outperforms GSAM in this measure. The reason for this is already visible in Fig.~\ref{fig:sam_pseudolabels}: if Grounding DINO proposes an imprecise bounding box, SAM segments the majority of the image as foreground. This severely affects the MAE value. An objective during optimizing the pseudo-label generation obviously is to detect and reject such failure cases.

The results of fine-tuning HitNet and SINet-V2 are presented in Table~\ref{tab:pseudo_label_comp} as well. Both models perform better when they are fine-tuned on the obviously suboptimal GSAM pseudo-labels. HitNet performs better than SINet-V2 independent of the pseudo-labels. This is why we will not consider SINet-V2 in the following experiments. Interestingly, looking at the $F_{\beta}^{w}$ measure, GSAM performs better on the CPD1K test dataset than both of the task-specific networks, which shows the strong zero-shot COD capabilities of SAM on the one hand and raises the question if COD models like HitNet are even needed on the other.

\begin{table}[ht]
    \centering
    \begin{tabular}{l|l|cc}
         Model & Pseudo-label generator & $MAE \downarrow$ & $F_{\beta}^{w} \uparrow$ \\
         \hline
         GSAM & - & 0.087 & \textbf{0.722} \\
         Baseline HitNet & - & \textbf{0.009} & 0.564 \\
         \hline
         HitNet & GSAM & 0.092 & \textbf{0.716} \\
         HitNet & Baseline HitNet & \textbf{0.008} &  0.624 \\
         SINet-V2 & GSAM & 0.064 & 0.619 \\
         SINet-V2 & Baseline HitNet & 0.009 & 0.599 \\
    \end{tabular}
    \caption{Results of the two pseudo-label generator approaches indicated by their zero-shot transfer performance and the two state-of-the-art COD models HitNet and SINet-V2 when fine-tuned on all pseudo-labels (i.e. not frugally learned), respectively, without any pseudo-label optimization.}
    \label{tab:pseudo_label_comp}
\end{table}

\begin{figure}[htbp]
    \centering
    \begin{subfigure}{\textwidth}
        \centering
        \includegraphics[width=\textwidth]{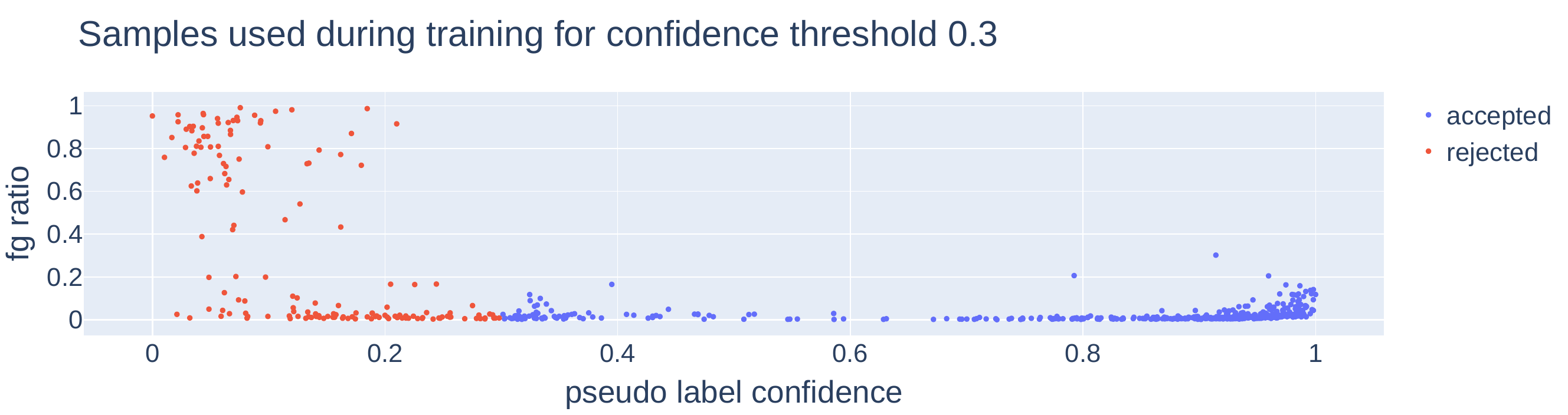}
        \caption{\textbf{Confidence thresholding:} Pseudo-labels with a confidence score larger than threshold $t_c$ are used during training, while all other pseudo-labels are rejected. The empirically optimized threshold for the CPD1K dataset is $t_c=0.3$.}
        \label{fig:sam_conf_threshold}
    \end{subfigure}
    \begin{subfigure}{\textwidth}
        \centering
        \includegraphics[width=\textwidth]{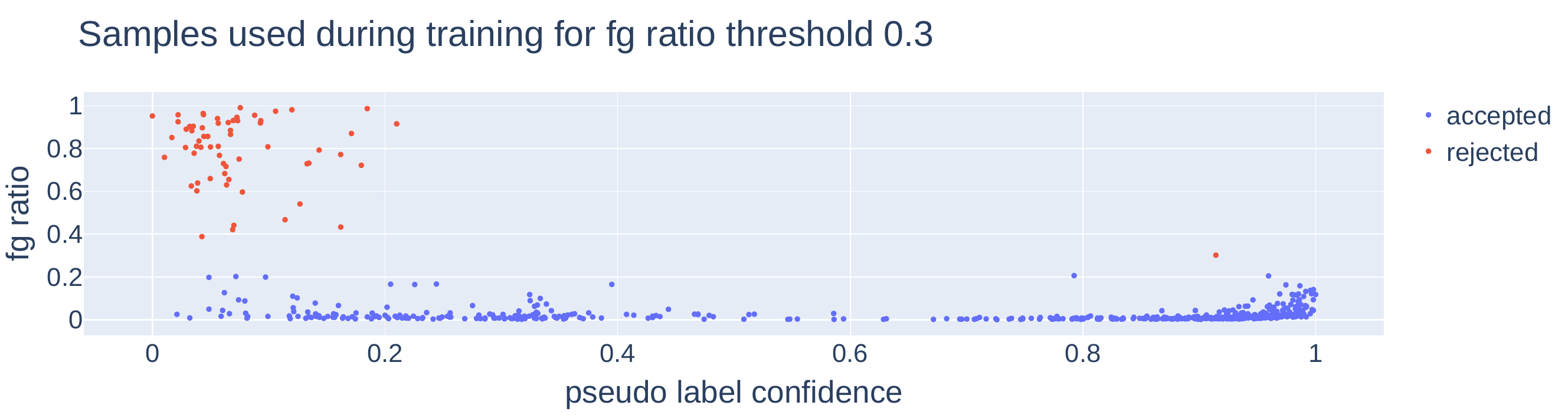}
        \caption{\textbf{Foreground ratio thresholding:} Pseudo-labels with a foreground ratio smaller than threshold $t_f$ are used during training, while all other pseudo-labels are rejected. The empirically optimized threshold for the CPD1K dataset is $t_f=0.3$. As the performance is similar to the confidence thresholding in Fig.~\ref{fig:sam_conf_threshold}, but foreground ratio thresholding can be error-prone due to the assumption of a small object size, we do not consider this approach for the following experiments.}
        \label{fig:sam_fg_threshold}
    \end{subfigure}
    \begin{subfigure}{\textwidth}
        \centering
        \includegraphics[width=\textwidth]{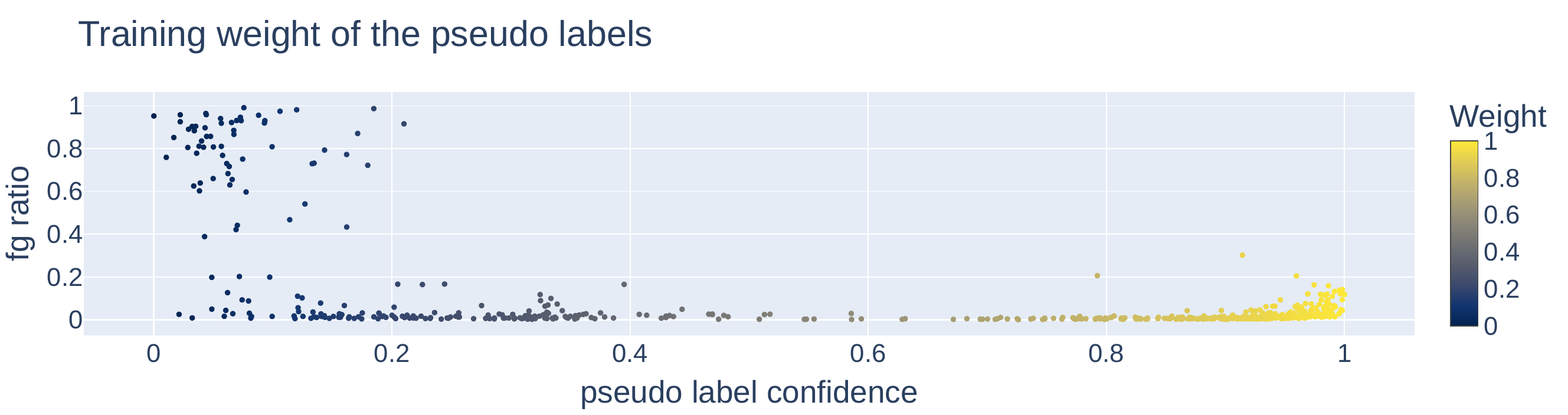}
        \caption{\textbf{Loss re-weighting:} The pseudo-labels found by GSAM in the CPD1K dataset are sorted by their confidence score. This confidence value is used to weight the pseudo-label-based training samples using loss re-weighting. The yellow color indicates a high confidence score leading to a higher weight during training.}
        \label{fig:sam_weights}
    \end{subfigure}
    \caption{Qualitative evaluation of the three different strategies proposed for the optimization of GSAM-based pseudo-label generation. Confidence thresholding and foreground ratio thresholding are complementing each other. Since confidence thresholding provides better generalization, we discard foreground ratio thresholding at this point.}
    \label{fig:pseudo_label_strategies}
\end{figure}

\noindent\textbf{Optimization of GSAM-based pseudo-label generation:} Now that a baseline for models trained on pseudo-labels is established, a closer look at the quality of the pseudo-labels is needed. We want to evaluate the confidences Grounding DINO provides for its bounding box proposals and the foreground ratio introduced in Eq.~\ref{eq:fg_ratio}. We assume that a sample with low confidence or high foreground ratio is likely to be a False Positive (FP) and leads to a reduced model performance. Fig.~\ref{fig:pseudo_label_strategies} shows a qualitative assessment of the three different optimizations for the pseudo-label generation as presented in Section~\ref{subsec:learning_with_noisy_labels}.
\label{sec:optimization_of_pseudo_label}

\begin{itemize}
\item \textbf{Confidence thresholding:} The first approach uses the confidence values provided by Grounding DINO to filter the training samples following Eq.~\ref{eq:conf_thresh}. Different confidence thresholds are used to determine the influence of imprecise pseudo-labels in the training process. We determined an optimal threshold of 0.3 for the CPD1K dataset. The samples accepted (blue) and rejected (red) during training are shown in Fig.~\ref{fig:sam_conf_threshold}.

\item \textbf{Foreground ratio thresholding:} The second approach uses the foreground ratio $ratio_{fg}$ according to Eq.~\ref{eq:fg_ratio} to filter the pseudo-labels for training. This approach can be error-prone as the size of a camouflaged object compared to the image is assumed to be small. The heuristic fails if this assumption is violated. As we did not see any advantage of this approach compared to the confidence thresholding, we only consider confidence thresholding for the following experiments. The pseudo-labels accepted for training are visualized in Fig.~\ref{fig:sam_fg_threshold}.

\item \textbf{Loss re-weighting:} The third approach uses the confidences of Grounding DINO to re-weight the loss of the corresponding sample as defined in Eq.~\ref{eq:loss_reweighting}. This means that the sample with the lowest confidence has basically no weight during training. The higher the confidence, the higher the sample's weight. This should lead to the model learning more from rather confident pseudo-labels according to Grounding DINO. Fig.~\ref{fig:sam_weights} shows a visualization of this re-weighting approach for all pseudo-labels found by GSAM in the CPD1K dataset.
\end{itemize}

Table~\ref{tab:pseudo_label_opt} provides an ablation study on the effectiveness of confidence thresholding and loss re-weighting. Both approaches together are slightly enhancing the $F_{\beta}^{w}$ measure, but severely improve the MAE value. This means that we can successfully filter out pseudo-labels, where large background regions are incorrectly segmented as foreground. In this experiment, we use all pseudo-labels found by GSAM in the CPD1K dataset.

\begin{table}[ht]
    \centering
    \begin{tabular}{l|c|c|c|cc}
         Model & Pseudo-label generator & Confidence thresholding (0.3) & Loss re-weighting & $MAE \downarrow$ & $F_{\beta}^{w} \uparrow$ \\
         \hline
         \multirow{4}{*}{HitNet} & \multirow{4}{*}{GSAM} & \xmark & \xmark & 0.023 & 0.633 \\
          & & \xmark & \cmark & 0.024 & 0.648 \\
          & & \cmark & \xmark & \textbf{0.005} & 0.732 \\
          & & \cmark & \cmark & \textbf{0.005} & \textbf{0.738}
    \end{tabular}
    \caption{Ablation study on the effect of confidence thresholding and loss re-weighting on the pseudo-label quality and thus the quality of the resulting HitNet model after training. In this experiment, we use $k=30$ pseudo-labels found by GSAM in the CPD1K training dataset and average over $i=10$ runs. We evaluate on CPD1K test.}
    \label{tab:pseudo_label_opt}
\end{table}


\subsection{Comparison between fully supervised learning and self-supervised frugal learning}
\label{sec:experiments_comp}

\noindent\textbf{Experimental setup:} The setup used for the experiments is largely the same as in Section~\ref{subsec:camouflaged_frugal_learning}. We use $k=30$ randomly picked samples for training and we perform ten runs to average the evaluation measures and to compensate for any random effects. Furthermore, only HitNet will be used in the following experiments since HitNet outperformed SINet-V2 in all experiments conducted before.

\noindent\textbf{Results:} Table~\ref{tab:results_comparison} shows the results. We consider all four evaluation measures to follow the standard COD evaluation protocols~\cite{Fan.b} in this summarizing table. The fully supervised HitNet model trained on the entire CPD1K training dataset with the GT labels serves as baseline in the first row. The remainder of the table shows the frugally learned models with only 30 samples, which is about 6\,\% of the full training dataset. The fully supervised but frugally learned HitNet model shows a relative performance drop of about 10\,\% compared to the baseline model according to the $F_{\beta}^{w}$ measure. We then show two variants of the a self-supervised frugally learned HitNet, where the pseudo-labels are generated by the original GSAM and the proposed optimized GSAM. Our optimized GSAM-based HitNet outperforms its original GSAM-based counterpart and it also slightly outperforms the HitNet model trained on the human-labeled GT annotations. We consider those results as highly promising to train camouflaged human detection frugally and in a self-supervised manner.

\begin{table}[ht]
    \centering
    \begin{tabular}{l|l|l|l|cccc}
         Model & Learning strategy & Label source & Training samples & $MAE \downarrow$ & $S \uparrow$ & $E_{\phi} \uparrow$ & $F_{\beta}^{w} \uparrow$ \\
             \hline
    HitNet & fully supervised & GT labels & 480 (full) & 0.003 & 0.900 & 0.962 & 0.828 \\
    \hline
         \multirow{3}{*}{HitNet} & fully supervised & GT labels & \multirow{3}{*}{30 (frugal)} & \textbf{0.005} & 0.852 & 0.936 & 0.734  \\
          & self-supervised & GSAM & & 0.023 & 0.812 & 0.843 & 0.633 \\
          & self-supervised & Optimized GSAM & & \textbf{0.005} & \textbf{0.855} & \textbf{0.939} & \textbf{0.738} \\
    \end{tabular}
    \caption{Comparison of the fully supervised HitNet model trained on the entire CPD1K training dataset with frugally learned variants of HitNet. Our proposed optimized GSAM generates pseudo-labels that achieve on par performance compared to the same model trained frugally on teh GT labels. With only about 6\,\% of the training data, we achieve a relative performance drop of only about 10\,\% according to the $F_{\beta}^{w}$ measure.}
    \label{tab:results_comparison}
\end{table}

\subsection{Limitations of Grounded SAM for COD}
We have seen in Table~\ref{tab:pseudo_label_comp} that GSAM alone can achieve comparable performance compared to HitNet trained on pseudo-labels considering the $F_{\beta}^{w}$ measure. However, there is a bias of the CPD1K dataset that becomes highly relevant here: each image always contains exactly one human. This is an assumption, which does not hold in the real world. Most images are expected to come without objects. We now evaluate the performance of GSAM and our proposed fine-tuned HitNet under such a setting. For that we will use the background images of the COD10K dataset. Please note that both the pre-trained and the fine-tuned HitNet have never seen those background images during training as they are not part of the regular COD10K training dataset.

\begin{figure}[htbp]
    \centering
    \begin{subfigure}{0.24\textwidth}
        \centering
        \includegraphics[width=\textwidth]{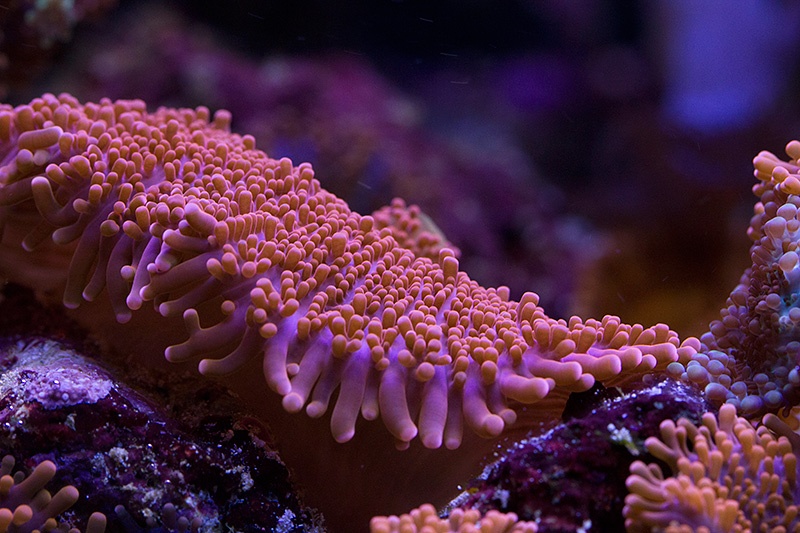}
    \end{subfigure}
    \begin{subfigure}{0.24\textwidth}
        \centering
        \includegraphics[width=\textwidth]{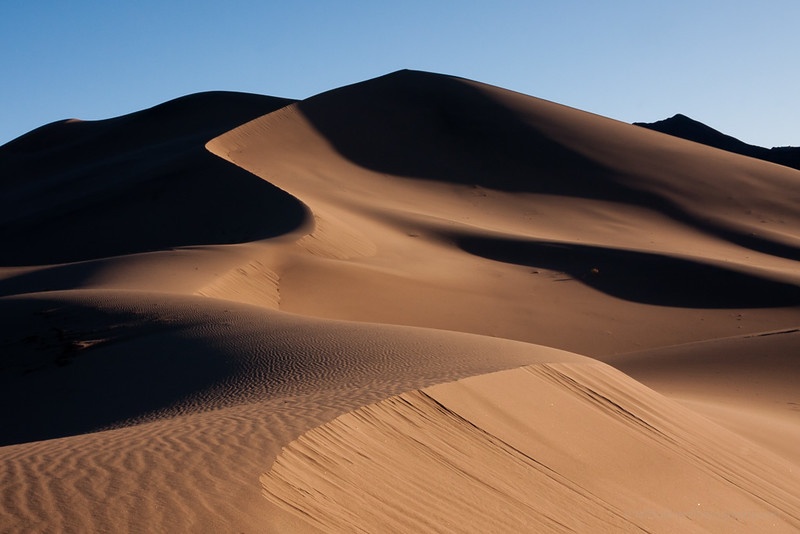}
    \end{subfigure}
    \begin{subfigure}{0.24\textwidth}
        \centering
        \includegraphics[width=\textwidth]{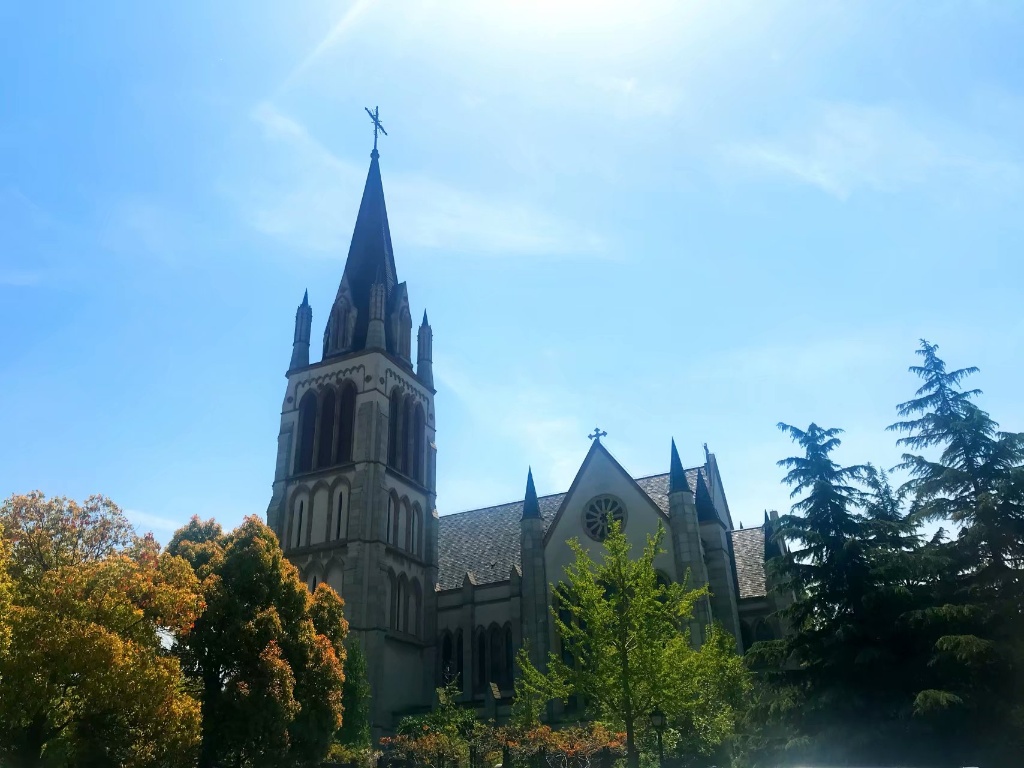}
    \end{subfigure}
    \begin{subfigure}{0.24\textwidth}
        \centering
        \includegraphics[width=\textwidth]{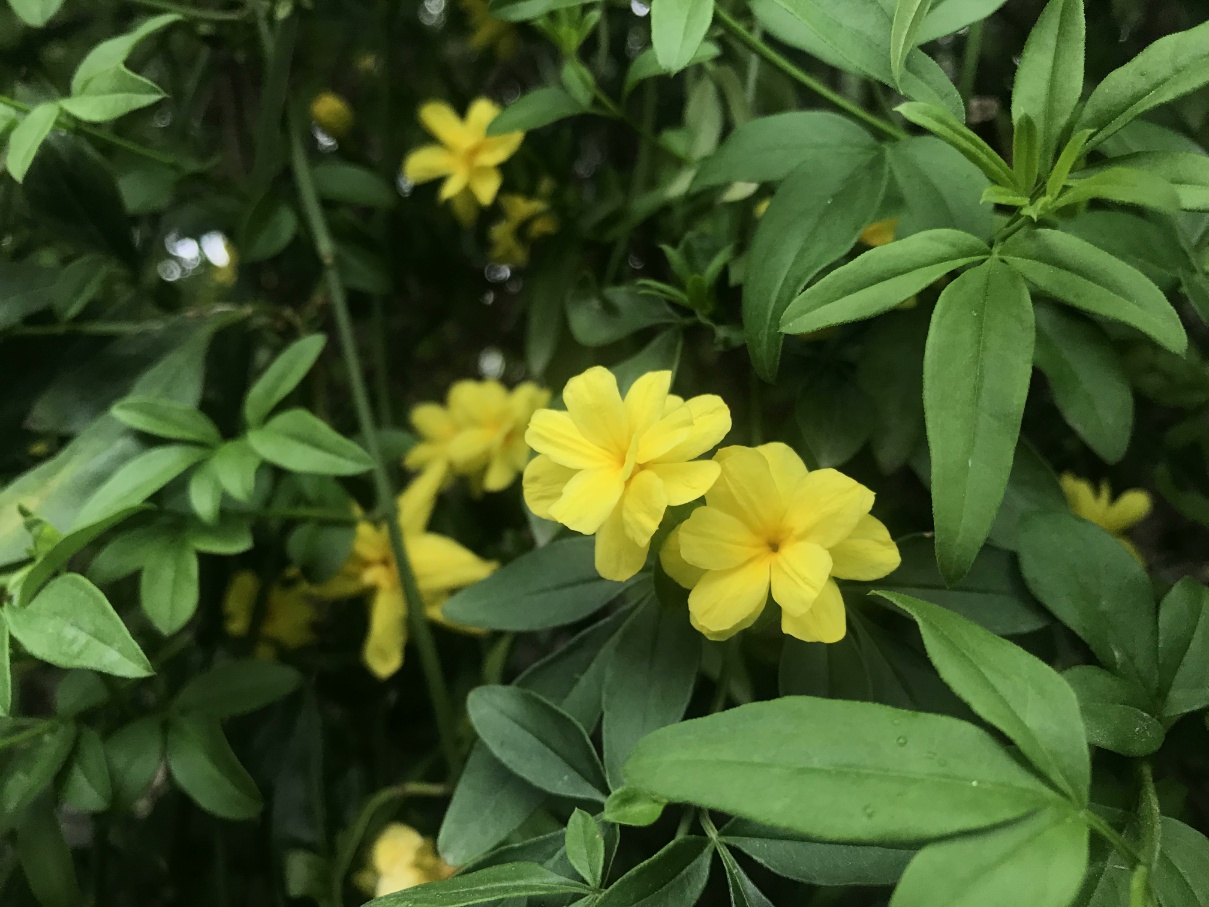}
    \end{subfigure}
    
    \caption{Background images taken from the COD10K dataset.}
    \label{fig:background}
\end{figure}

\noindent\textbf{Dataset:} As mentioned before, we will use the background images of COD10K for this experiment. This dataset contains 1068 images from different scenes like underwater, desert, sky, and vegetation. Example images of the background dataset can be seen in Fig.~\ref{fig:background}.

\noindent\textbf{Evaluation measures:} Since we evaluate background images only without any objects, we have to use evaluation measures that do not consider true positive (TP) or false negative (FN) detections as we will not have any TPs and thus no FNs in our data. There are two measures available: the False Positive Rate (FPR) calculates the probability of a false alarm and the True Negative Rate (TNR) calculates the probability to correctly identify negative pixels. The formula for the FPR and TNR can be seen in Eq.~\ref{eq:fpr} and Eq.~\ref{eq:tnr}, respectively.

\begin{equation}
    FPR = \frac{FP}{TN + FP}
    \label{eq:fpr}
\end{equation}

\begin{equation}
    TNR = \frac{TN}{TN + FP}
    \label{eq:tnr}
\end{equation}

\noindent\textbf{Experimental setup:} The two HitNet models that we use here are (1)~the model trained on CPD1K under a fully fine-tuned setting using the original GT labels, and (2)~the model trained using the pseudo-labels generated by GSAM. We use the confidence thresholding approach rejecting all pseudo-labels with a confidence lower 0.3 and we use $k=30$ and $i=10$. For GSAM as competitor, we use the same settings as for generating pseudo-labels in Section~\ref{sec:experiments_ssl}.

\noindent\textbf{Results:} Table~\ref{tab:res_background_images} shows the results of the experiment. There is almost no difference between the HitNet model trained with a fully supervised learning strategy and the HitNet model trained using pseudo-labels. GSAM, however, is outperformed by a large margin. On most background images GSAM segments a large portion of the image as foreground, which results in a high FPR and low TNR score. Even though the performance of GSAM and HitNet are similar when using images with exactly one object, HitNet significantly outperforms GSAM when this assumption does not hold anymore. This shows that GSAM alone is not suitable for solving the COD task in a realistic setting.

\begin{table}[ht]
    \centering
    \begin{tabular}{lcc}
         Model & FPR $\downarrow$ & TNR $\uparrow$ \\
         \hline
         HitNet fully supervised & 0.017 & 0.982 \\
         HitNet pseudo-labels & 0.020 & 0.979 \\
         GSAM & 0.680 & 0.319
    \end{tabular}
    \caption{Results on pure background images without any objects in terms of FPR and TNR. Both HitNet variants outperform GSAM by a large margin. The reasons is that GSAM prompted by the phrase 'soldier' seems to be biased towards finding any camouflaged humans. This shows that GSAM alone is not capable of performing zero-shot COD.}
    \label{tab:res_background_images}
\end{table}


\section{CONCLUSION}

\label{sec:conclusion}
We show that the challenging task of camouflaged human detection can be tackled without requiring large amounts of data or any costly, human-labeled data by leveraging frugal learning techniques. 
Our proposed frugal learning approach using GSAM allows us to train a HitNet with only 30 samples but no GT labels, achieving the same performance as a HitNet trained with 30 samples and human-labeled GT.
Overall, our frugally trained model is just 10\,\% worse than a model trained in a fully supervised way (using all available images with human-labeled GT).
However, our experiments also show that the strong zero-shot transfer performance of GSAM only holds when every image contains at least one object. Under a more realistic setting that includes empty background images, GSAM's performance significantly deteriorates whereas the fully supervised HitNet and our frugally trained HitNet show robustness to this setting. 
Since our approach does not require any additional manual effort, scaling to much larger unlabeled datasets seems highly promising.
Additionally, other foundation models and pseudo-label generators should be investigated and compared.
In this context, investigating ways to improve the performance of GSAM, especially to handle empty background images, could also provide meaningful insights and guide future researchers to solve the task of COD without costly human labels.

\bibliography{main} 
\bibliographystyle{spiebib} 

\end{document}